\documentclass[runningheads]{llncs}
\usepackage{graphicx}

\usepackage{tikz}
\usepackage{hyperref}
\usepackage{comment}
\usepackage{amsmath,amssymb} 
\usepackage{color}

\usepackage[accsupp]{axessibility}  

\usepackage{booktabs}
\usepackage{slashbox}
\usepackage{subcaption}
\usepackage{pifont}

\usepackage{multirow}
\usepackage{makecell}
\usepackage{boldline}
\usepackage{enumitem}
\usepackage{xcolor}
\setcellgapes{3pt}
\usepackage[symbol]{footmisc}

\DeclareMathOperator*{\argmax}{arg\,max}
\usepackage[misc]{ifsym}
\usepackage{wrapfig}

\usepackage{xspace}
\makeatletter
\DeclareRobustCommand\onedot{\futurelet\@let@token\@onedot}
\def\@onedot{\ifx\@let@token.\else.\null\fi\xspace}

\def\eg{\emph{e.g}\onedot} 
\def\ie{\emph{i.e}\onedot} 
 
\def\etc{\emph{etc}\onedot} 
 
\def\etal{\emph{et al}\onedot}
\makeatother

\begin{document}
\pagestyle{headings}
\mainmatter
\def\ECCVSubNumber{0300}  

\title{Active Pointly-Supervised \\Instance Segmentation} 

\titlerunning{Active Pointly-Supervised Instance Segmentation}
%
\author{
Chufeng Tang\inst{1} \and
Lingxi Xie\inst{2} \and
Gang Zhang\inst{1} \and
Xiaopeng Zhang\inst{2} \and \\
Qi Tian\inst{2}\textsuperscript{(\Letter)} \and
Xiaolin Hu\inst{1,3,4}\textsuperscript{(\Letter)}
}
\authorrunning{C. Tang et al.}
%
\institute{
$^1$Department of Computer Science and Technology, Institute for AI, BNRist,\\ State Key Laboratory of Intelligent Technology and Systems, Tsinghua University\\
$^2$Huawei Inc. \qquad $^3$Chinese Institute for Brain Research (CIBR) \\
$^4$IDG/McGovern Institute for Brain Research, Tsinghua University \\
\email{\{tcf18, zhang-g19\}@mails.tsinghua.edu.cn},
\email{\{198808xc, zxphistory\}@gmail.com} \\
\email{tian.qi1@huawei.com}, \enspace \email{xlhu@mail.tsinghua.edu.cn}
}
\maketitle

\begin{abstract}
  The requirement of expensive annotations is a major burden for training a well-performed instance segmentation model. In this paper, we present an economic active learning setting, named active pointly-supervised instance segmentation (APIS), which starts with box-level annotations and iteratively samples a point within the box and asks if it falls on the object. The key of APIS is to find the most desirable points to maximize the segmentation accuracy with limited annotation budgets. We formulate this setting and propose several uncertainty-based sampling strategies. The model developed with these strategies yields consistent performance gain on the challenging MS-COCO dataset, compared against other learning strategies. The results suggest that APIS, integrating the advantages of active learning and point-based supervision, is an effective learning paradigm for label-efficient instance segmentation.
\end{abstract}

\section{Introduction} \label{sec:introduction}
Instance segmentation aims to predict a pixel-wise mask with a category label for each instance in the given image. Despite the rapid development of the instance segmentation methods, the requirement of a massive amount of labeled data is still a heavy burden of training a well-performed instance segmentation model. For example, the annotation of a polygon-based mask for an object in MS-COCO requires 79.2 seconds~\cite{mscoco} on average and even higher for the more precise mask annotations in LVIS~\cite{lvis}, which is considerably more time-consuming than annotating a bounding box (\eg, 7 seconds via clicking extreme points~\cite{papadopoulos2017extreme}).

\begin{figure}[t]
\begin{center}
  \includegraphics[width=0.98\linewidth]{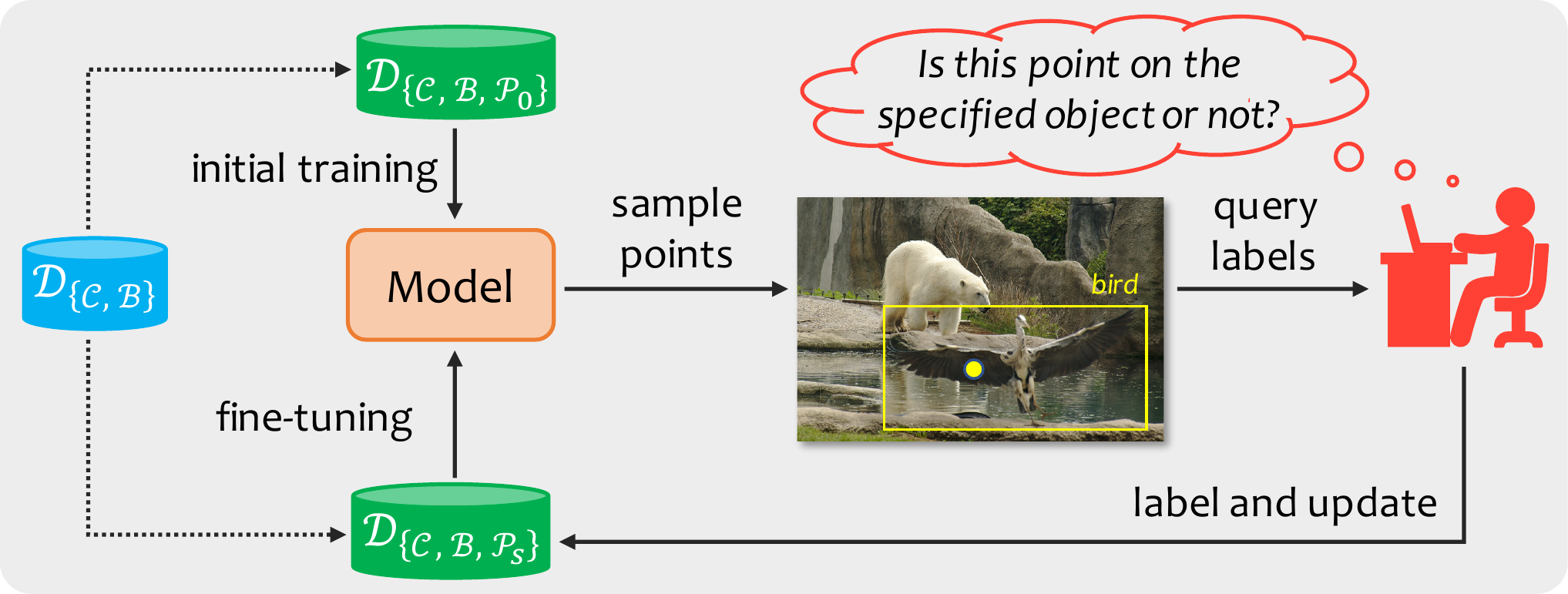}
\end{center}
  \caption{
    Overview of the training pipeline for the proposed APIS setting where the annotator is asked for labeling whether the point falls on the specified object (\ie, the bird) or not. The expected label in this case is `yes'. $\mathcal{D}$ is the training data and $\mathcal{C},\mathcal{B},\mathcal{P}$ are the category, bounding box, and point annotations, respectively (see Section~\ref{sec:formulation} for details).
  }
  \label{fig:fig1}
\end{figure}

In general, weak supervision and active learning are usually two effective ways to reduce the annotation cost. For the task of instance segmentation, a number of existing works attempted to predict instance masks with weak supervisions, such as category tags~\cite{arun2020weakly,pont2016multiscale,zhu2019learning}, bounding boxes~\cite{hsu2019weakly,khoreva2017simple,lan2021discobox,boxinst}, and points~\cite{pointsup,laradji2020proposal}. However, active learning for instance segmentation has been less investigated. Wang \etal~\cite{wang2020semi} first explored the possibility on medical image analysis, but, to the best of our knowledge, no existing work has studied this setting on more complex and larger-scale datasets (\eg, MS-COCO~\cite{mscoco}).

In this paper, we present a new setting named \textbf{active pointly-supervised instance segmentation (APIS)} to study active learning algorithms for instance segmentation with point supervisions. Under the proposed setting where each image in the data pool has been annotated with category labels and bounding boxes, the goal of the algorithms is selecting the most informative \textit{points} for labeling to maximize the model's performance. Fig.~\ref{fig:fig1} illustrates the training pipeline of APIS. Compared to the typical active learning settings that the most informative \textit{images} or \textit{instances} are selected and annotated with boxes and masks, APIS can be studied in a more fine-grained manner because it allocates annotation budgets to \textit{pixels}, and the annotation of points is considerably faster and cheaper. As stated in a previous work~\cite{pointsup}, labeling whether a point falls on the specified object or not takes only $0.9$ seconds on average. Note that active learning is a training strategy that point labels are only queried during the training phase of APIS, while no annotations are provided for testing.

APIS raises an important problem that has not been studied before, \ie, \textit{how to estimate the informativeness of a point}, where the informativeness can be roughly defined as the potential gain of segmentation accuracy (\eg, in terms of mAP) if this point is annotated. However, estimating the precise accuracy gain for each point is obviously infeasible. In the literature, uncertainty is widely used to estimate the informativeness of an example. Inspired by this, we designed several metrics to estimate the uncertainty of a point based on the model's prediction and select the most uncertain point of each instance for annotation. The labeled points are used together with the category and box annotation to update (\eg, fine-tuning) the instance segmentation model.

Extensive experiments on the challenging MS-COCO dataset demonstrate that the model trained with the actively acquired points performed consistently better than the model trained with randomly sampled points during each active learning step. \textit{Especially, we found that entropy, a conceptually simple metric, works the best among all the proposed metrics for uncertainty estimation.} To understand the results of APIS, we went deep into the point selection process and studied the training dynamics, point distribution and point difficulties. The analyses reveal that the proposed sampling metric provides a good estimation of the point informativeness and therefore leads to higher performance. In addition to the random sampling baseline, we further compared APIS against a setting of active instance segmentation with full supervisions, where the most informative images or instances are selected for mask annotation. Conditioned on the same annotation budget and training time, the model developed under the APIS setting outperformed all other competitors. \textit{The results suggest that active learning cooperates effectively with point supervisions, which can further boost the instance segmentation performance under a limited annotation budget.} We hope the promising results, as well as the comprehensive analysis presented in this work, will draw the attention of the community to APIS and other label-efficient visual recognition techniques.

The contribution of this work is summarized as follows:
\begin{itemize} 
  \item We present a new active instance segmentation setting APIS where the goal is to sample the most informative points to maximize the  model's performance. To our knowledge, this work is the first to explore active learning for instance segmentation with point supervisions.
  \item We estimate the informativeness with the uncertainty of point predictions, and the model trained with the actively acquired points consistently outperformed the random sampling counterparts on MS-COCO.
  \item We provide comprehensive comparisons and analyses to understand the results of APIS and concluded that APIS successfully combines the advantages of active learning and point-based supervision to reduce the annotation burden of instance segmentation.
\end{itemize}

\section{Related Work} \label{sec:related}
\noindent \textbf{Instance Segmentation}. Currently, the fully supervised methods still dominate the popular instance segmentation benchmarks~\cite{cityscapes,lvis,mscoco}.  Mask R-CNN~\cite{maskrcnn} and its follow-up methods~\cite{htc,huang2019mask,liu2018path} predict masks based on the region-level features. One-stage methods like CondInst~\cite{condinst} and SOLO~\cite{wang2021solo} directly segment instances at image-level by learning instance-aware kernels. Recently, some query-based methods~\cite{cheng2021masked,solq,fang2021instances} further boost the segmentation performance. For the weakly supervised paradigm, category tags~\cite{arun2020weakly,pont2016multiscale,zhu2019learning} are the simplest supervision but the results are usually uncompetitive. With bounding box annotations, the recently proposed BoxInst~\cite{boxinst} and DiscoBox~\cite{lan2021discobox} significantly outperformed previous methods~\cite{hsu2019weakly,khoreva2017simple} on MS-COCO. In addition, PointSup~\cite{pointsup} further reduced the gap to the fully-supervised methods by training with boxes and several randomly sampled points for each instance, which is similar the random sampling baseline in APIS where the points are accessed for training step by step. However, there are still great differences that APIS focus on active learning for instance segmentation while PointSup~\cite{pointsup} is for weakly-supervised learning.

\noindent\textbf{Active Learning}. Over the past decades, a great number of active learning algorithms have been proposed but mostly designed for image classification, which can be roughly divided into two categories: uncertainty-based~\cite{beluch2018power,gal2017deep,joshi2009multi,Lewis1994HeterogeneousUS,wang2016cost} and diversity-based~\cite{agarwal2020contextual,hasan2015context,sener2017active,sinha2019variational} algorithms. In recent years, some researchers shifted their attention to the downstream visual recognition tasks such as object detection~\cite{aghdam2019active,choi2021active,haussmann2020scalable,kao2018localization,liu2021influence,roy2018deep,yuan2021multiple} and semantic segmentation~\cite{Shin_2021_ICCV,wu2021redal,yang2017suggestive}. However, for the task of instance segmentation, the potential of active learning has been less explored. Only one published work~\cite{wang2020semi} preliminarily studied this problem on the medical image datasets where a triplet uncertainty metric was calculated with the predicted mask IoU scores of Mask Scoring R-CNN~\cite{huang2019mask}, which makes the metric relying on a specialized architecture. By contrast, we first studied this problem on the most commonly used MS-COCO dataset and the proposed metrics are model-agnostic. Furthermore, different from all existing settings where images or instances~\cite{desai2020towards} are selected for labeling, the proposed APIS setting performs in a fine-grained and weakly supervised manner, \ie, selecting and labeling points. A few works~\cite{desai2019adaptive,pardo2021baod} studied weak supervisions for active object detection while they focused on how to decide the annotation scheme (strongly or weakly) for an image, which are complementary to APIS.

\noindent\textbf{Point-Based or Click-Based Segmentation}. Point-based supervision has been studied in various image segmentation tasks, including semantic segmentation~\cite{bearman2016s,qian2019weakly}, instance segmentation~\cite{pointsup,laradji2020proposal}, and panoptic segmentation~\cite{li2021fully}. In addition, point clicks are widely used in interactive annotation~\cite{benenson2019large,Jang2019InteractiveIS,Li_2018_CVPR,Maninis_2018_CVPR,Xu_2016_CVPR} methods, which usually requires mask annotations for training the model. During testing (\ie, annotating an unseen image), the user is asked to provide some corrective point clicks iteratively. APIS is intrinsically different to these methods that the model only interacts with users during training and no additional labels are required during testing.

\section{Method} \label{sec:method}
\subsection{Problem Formulation} \label{sec:formulation}
In this section, we formally define the proposed active pointly-supervised instance segmentation (APIS) setting. Suppose we collect a large training dataset of $N$ images denoted as $\mathcal{D}=\{\mathcal{I}_i\}{_{i=1}^N}$ with annotations $\{\mathcal{C},\mathcal{B}\}$, where $\mathcal{C}=\{\mathcal{C}_i\}{_{i=1}^N}$ is the category annotation, $\mathcal{B}=\{\mathcal{B}_i\}{_{i=1}^N}$ is the box annotation, and $Q_i=|\mathcal{B}_i|=|\mathcal{C}_i|$ is the number of instances in the $i^\mathrm{th}$ image. We studied a typical setting of APIS that each instance is sampled with the same number of points and all the points should be located in the corresponding ground-truth bounding box. Additionally, we also studied a scenario where \textit{not} all instances were labeled with the same number of points, see Appendix~\ref{appendix_fewer} for details.

Before the active learning cycle starts, we randomly sample and annotate one point for each instance. The initial set of points is denoted as:
\begin{equation}
  \mathcal{P}_0=\{ (x_{ij}^{0}, y_{ij}^{0}, u_{ij}^{0}) \mid 1 \leq i \leq N, 1 \leq j \leq Q_i \},
\end{equation}
where $(x_{ij}^{0}, y_{ij}^{0})$ is the coordinates of a point that is located in the $j^\mathrm{th}$ bounding box of the $i^\mathrm{th}$ image, and $u_{ij}^{0} \in \{0,1\}$ is the point label which indicates whether the point falls on the foreground object or not. The size of $\mathcal{P}_0$ is $Q=\sum^{N}_{i=1}Q_{i}$ that equals to the total number of instances in $\mathcal{D}$. The instance segmentation model $\mathcal{M}_0$ is initialized by training with the above annotations $\{\mathcal{C},\mathcal{B},\mathcal{P}_0\}$.

At the $s^\mathrm{th}$ ($s\geq1$) active learning step, the informativeness of points is estimated based on the predictions of the previous model $\mathcal{M}_{s-1}$. We designed several uncertainty-based metrics to estimate the point informativeness, which will be explained in Section~\ref{sec:point_selection}. The most informative point of each instance is selected and the annotators are asked to label it. The newly labeled points are merged with $\mathcal{P}_{s-1}$ and formed the new point set $\mathcal{P}_s$:
\begin{equation}
  \mathcal{P}_s = \mathcal{P}_{s-1} \cup \{ (x_{ij}^{s}, y_{ij}^{s}, u_{ij}^{s}) \mid 1 \leq i \leq N, 1 \leq j \leq Q_i \},
\end{equation}
where the number of points in $\mathcal{P}_s$ is $(s+1) \times Q$. Subsequently, the model $\mathcal{M}_s$ is fine-tuned from $\mathcal{M}_{s-1}$ with all available annotations $\{\mathcal{C},\mathcal{B},\mathcal{P}_s\}$. The above process is repeated multiple steps until the annotation budget has been exhausted or a satisfactory performance has been achieved.

\subsection{Point Selection for APIS} \label{sec:point_selection}
In this section, we tackle the core problem raised by APIS that how to define a point's informativeness. Note that informativeness is not totally determined by correctness, \eg, if the model  predicts a point as positive with a high confidence while the ground-truth is negative, annotating it may not be an ideal solution (see Sec.~\ref{sec:analysis}). That said, even provided the mask labels, it is still difficult to determine which point has a potentially large contribution. We refer to some existing active learning methods~\cite{beluch2018power,gal2017deep,joshi2009multi,Lewis1994HeterogeneousUS,wang2016cost} to study uncertainty, and use the uncertainty of points to rank their informativeness.

In modern instance segmentation models, a ground-truth instance is usually assigned as the learning target for multiple predictions during training, known as label assignment~\cite{zhu2020autoassign}, which makes NMS (Non-Maximum Supression) being a necessary process during inference. Suppose there are $K$ mask predictions $\{\mathbf{m}^k\}{_{k=1}^K}$ matching to a given instance in $\mathcal{D}$, where $\mathbf{m}^k \in [0,1]^{H \times W}$ is an image-level probability matrix and $H \times W$ is the shape of the image. Note that for CondInst~\cite{condinst} the prediction is already image-level while for Mask R-CNN~\cite{maskrcnn} the region-level prediction should be transformed to image-level. Denote the element of $\mathbf{m}^k$ with coordinates $(x,y)$ as $p^k_{xy}$ that indicates the probability of a point located at $(x,y)$ falling on the object. We designed several metrics to estimate the uncertainty for a point based on its predictive probability of $p^k_{xy}$.

\textbf{(1) Entropy of the Averaged Predictions.}
The Shannon Entropy~\cite{shannon2001mathematical} metric is commonly used in existing active learning algorithms~\cite{aghdam2019active,gal2017deep,haussmann2020scalable,wang2016cost}. In our case, since mask prediction is a binary classification problem for points, the entropy metric can be defined as:
\begin{equation}
  \mathcal{H}(p) = - p \log p - (1 - p) \log (1-p),
\end{equation}
where $p$ is the probability of any point. We simply average multiple predictions on the same point to calculate the entropy value, and the point with the highest entropy value is selected for the corresponding instance:
\begin{equation}
  (\hat{x},\hat{y}) = \argmax_{(x,y)\in\Omega}\mathcal{H}(\bar{p}_{xy}), \quad \bar{p}_{xy} = \frac{1}{K}\sum^{K}_{k=1}p^k_{xy},
\end{equation}
where $(\hat{x},\hat{y})$ is the coordinates of the actively selected point for the given instance and $\Omega$ is the spatial constraint for the point candidates. In our setting, the selected points are expected to fall inside the ground-truth bounding boxes. The intuition behind the above sampling strategy is straightforward yet reasonable, \ie, \textit{the more the probability close to 0.5, the higher the entropy, and the more likely the point should be selected}. Designing the constraint $\Omega$ carefully or averaging multiple predictions with some adaptive weights can make the above strategy more sophisticated, however it is not the focus of this work.

\textbf{(2) Disagreement Among Multiple Predictions.}
The intuition behind the disagreement metric is that \textit{if multiple predictions for the same point varied significantly, the model should be highly uncertain about that point and we should select that point for labeling}. This idea can be traced back to the classical \textit{query by committee} paradigm~\cite{melville2004diverse}. A number of works implicitly or explicitly followed this idea to select samples actively. For example, in the task of active object detection, the offsets between the boxes generated from different SSD layers~\cite{roy2018deep}, or the IoU scores between the proposals and final detected boxes~\cite{kao2018localization} were used to measure the disagreement. In our case, we adopt the \textit{variance} across different predictions for a point to measure the disagreement and the point with the largest \textit{variance} is selected for the corresponding instance:
\begin{equation}
  \label{eq:variance}
  (\hat{x},\hat{y}) = \argmax_{(x,y)\in\Omega}\mathbb{V}(\mathbf{p}_{xy}),
\end{equation}
where $\mathbf{p}_{xy} \in \mathbb{R}^{K}$ is the probability vector at the location $(x,y)$ and $\mathbb{V}(\cdot)$ calculates the variance of it.

In addition to sampling metrics mentioned above, how to form the prediction set $\{\mathbf{m}^k\}{_{k=1}^K}$ for a given instance is an important problem. In this work, by considering the properties of the APIS setting, we propose and compare several solutions. (a) \textit{Multiple Anchors}: Since a ground-truth instance is usually assigned to multiple anchors (where anchors are the positive locations for CondInst~\cite{condinst}, or the positive proposals for Mask R-CNN~\cite{maskrcnn}, \etc) during training, we naturally have multiple predictions for one instance. (b) \textit{Multiple Models}: Another solution is training the same models multiple times with different initializations to get multiple predictions, which is usually called Deep Ensembles~\cite{beluch2018power,lakshminarayanan2017simple} in the literature. (c) \textit{Multiple Scales}: Training multiple models is computationally inefficient. An alternative is to forward the same model multiple times under different conditions. In our case, the model is usually trained with multi-scale inputs, thus we can forward the model with an individual scale each time to get multiple predictions. The concept of multi-scale here can be freely replaced by other types of data augmentations (\eg, flipping, rotation).

We calculate the entropy and disagreement metrics on the above prediction sets individually and obtain several different point sampling strategies. Note that the case of multiple anchors also appear in each model or each forward pass, and we simply concatenate them as a single prediction set. For the instance that was not predicted ($K=0$), we randomly sample a point for it. For the instance that has only one prediction ($K=1$), we only use entropy as the metric.

\subsection{Baseline for APIS} \label{sec:baseline}
Note that there is no existing work can be directly compared to APIS since the problem of active instance segmentation has been less investigated.
In addition to the \textit{random sampling} baseline, inspired by the existing works for active image classification or object detection where the full labels are usually queried, we create a baseline setting for comparison, named \textit{active fully-supervised instance segmentation (AFIS)}, where the mask annotations are queried during each active learning step. Note that the category and box annotations $\{\mathcal{C},\mathcal{B}\}$ are also provided in advance. We designed several sampling strategies under the AFIS setting, which will serve as the baselines for APIS. A straightforward strategy is selecting the most informative \textit{images} and labeling masks for all instances in the image. An alternative is selecting and annotating the most informative \textit{instances}, which is a fine-grained solution that not all instances in the image are labeled. We call them \textit{image-level} selection and \textit{instance-level} selection for AFIS. See Appendix~\ref{appendix_AFIS} for more description of AFIS.

Under the AFIS setting, we propose two metrics to define the informativeness of an image or an instance.
(a) \textit{Mean Entropy}: Inspired by the aforementioned uncertainty-based metrics of APIS, we defined the instance uncertainty as the mean entropy of all points inside the ground-truth bounding box, and the uncertainty of an image is defined as the mean uncertainty of all instances in that image. The most uncertain images and instances are selected for the image-level and instance-level cases, respectively.
(b) \textit{Detection Quality}: The quality of mask prediction is usually dependent on the detection quality. For an instance that was not accurately detected, the potential of the mask label, if provided, may not be fully utilized. Therefore, we propose to select the instances with the higher detection quality for labeling masks. Since the ground-truth boxes are provided, we can easily use the detection loss (\eg, GIoU Loss~\cite{rezatofighi2019generalized}) as the metric for instance selection. For the image-level selection, we calculate the average detection loss of all instances for an image, and the images with the lowest loss are selected.

\section{Experiments} \label{sec:experiments}
\subsection{Experimental Settings}
We report results on the MS-COCO~\cite{mscoco} dataset. All the active selection strategies were applied on the \texttt{train2017} split, including 118k images with 860k instances, where the annotation for points was simulated by adopting the labels of the ground-truth instance masks on the corresponding location. All models were evaluated on the \texttt{val2017} split.

\textbf{Implementation Details.}
We mainly took CondInst~\cite{condinst} with ResNet-50 as the instance segmentation model and adopted the \texttt{AdelaiDet}~\cite{tian2019adelaidet} codebase. To train the model with point supervision, we followed the same training protocol as PointSup~\cite{pointsup} where the point prediction was sampled from the prediction map using bilinear interpolation and the per-pixel cross-entropy loss was calculated on the labeled points only. The initial model $\mathcal{M}_0$ was trained for 90k iterations with the initial learning rate being $0.01$ and decayed at iteration 60k and 80k, respectively. Other training settings were the same as CondInst. After that, the active learning process was repeated multiple times. At the $s^\mathrm{th}$ active learning step, $(s+1)$ points were labeled in total for each instance where $s$ points were actively acquired, denoted as $\mathcal{P}_s$ in the following experiments. By default, at each step, the model was fine-tuned for 30k iterations with the initial learning rate being $0.01$ and decayed every 10k iterations. All models were trained with the SGD optimizer and multi-scale data augmentation with mini-batch of 16 images on 8 NVIDIA Tesla V100 GPUs. The results were measured by the mask mAP(\%) metric of instance segmentation. 

\subsection{Design Principles of APIS}

\begin{figure}[t]
  \centering
  \begin{subfigure}[b]{0.49\textwidth}
    {\includegraphics[width=\textwidth]{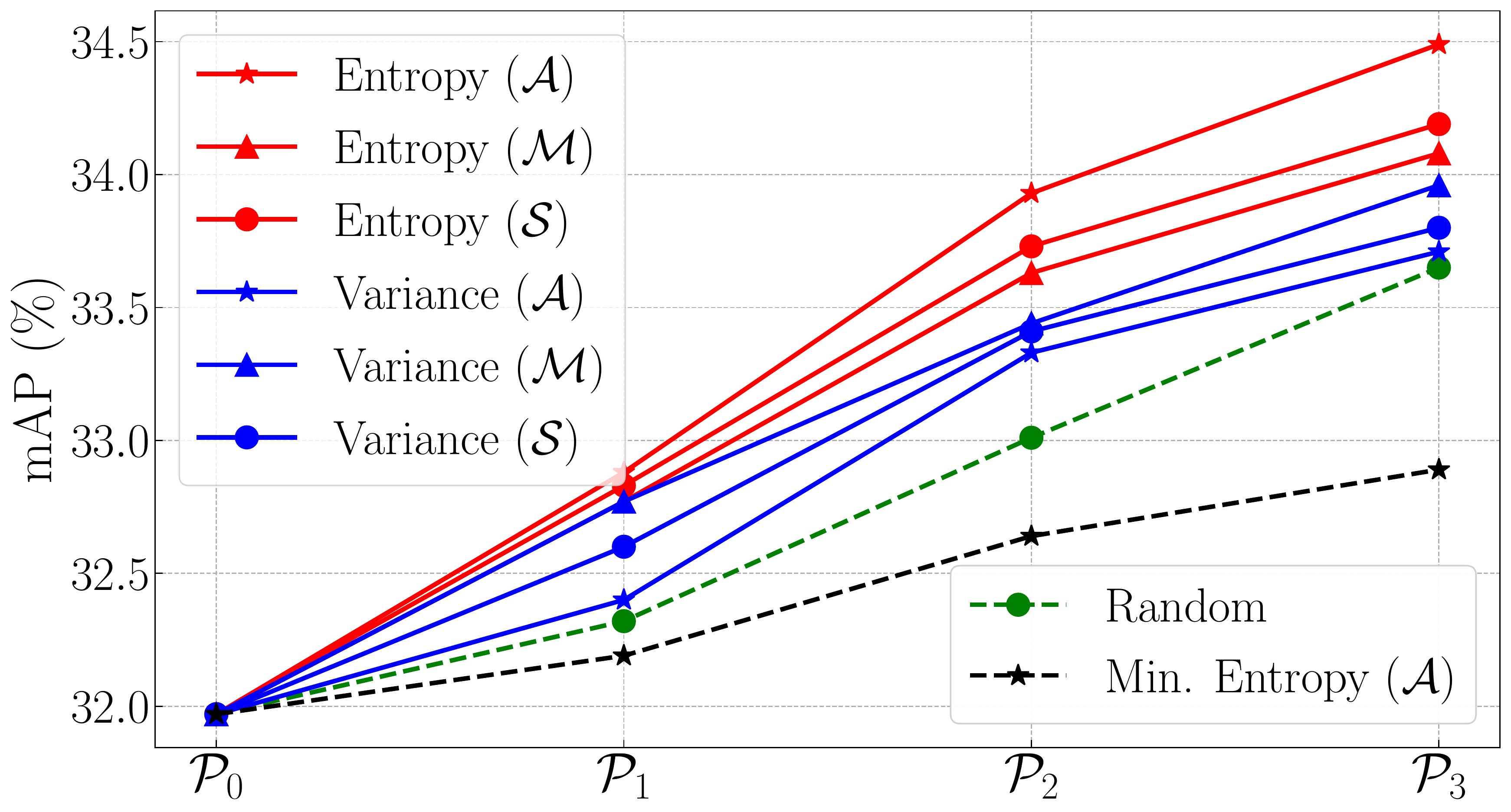}}
    \caption{Point selection strategies}
    \label{fig:ablation_selection_strategy}
  \end{subfigure}
  \hfill
  \begin{subfigure}[b]{0.49\textwidth}
    {\includegraphics[width=\textwidth]{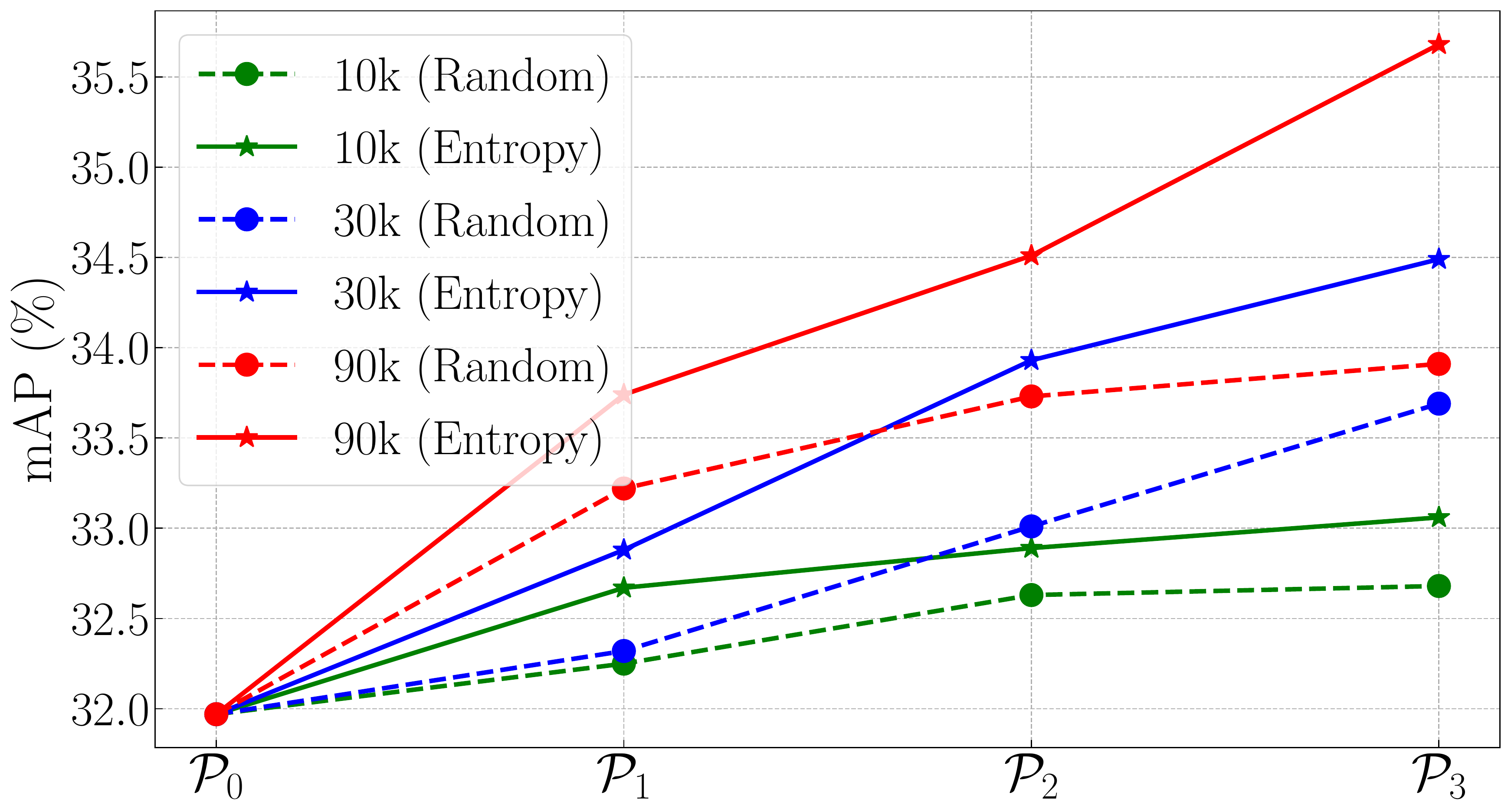}}
    \caption{Fine-tuning schedules}
    \label{fig:ablation_training_schedule}
  \end{subfigure} \\
  \caption{(a) Comparison of different point selection strategies. $\mathcal{A}$, $\mathcal{M}$ and $\mathcal{S}$ indicate the prediction sets constructed from multiple anchors, multiple models, and multiple scales, respectively. (b) Comparison of different fine-tuning schedules.}
  \label{fig:ablation}
\end{figure}

\textbf{Point Selection Strategy.} As introduced in Section~\ref{sec:point_selection}, we propose two metrics, named \textit{Entropy} and \textit{Variance} (\ie, the disagreement) here, to estimate the point uncertainty based on the prediction set, and there are three types of prediction sets: \textit{Multiple-Anchors} ($\mathcal{A}$), \textit{Multiple-Models} ($\mathcal{M}$), and \textit{Multiple-Scales} ($\mathcal{S}$). We composed them into six different point selection strategies, and we observed that all these six strategies performed consistently better than the random sampling baseline, as shown in Fig.~\ref{fig:ablation_selection_strategy}. The results demonstrate the effectiveness of active point selection. In addition, the strategies that taking \textit{Entropy} as the metric usually performed better than the \textit{Variance} counterparts. When using \textit{Entropy} as the metric, constructing the prediction set from multiple anchors ($\mathcal{A}$) performed better than from multiple models ($\mathcal{M}$) or scales ($\mathcal{S}$), and the latter two solutions are obviously inefficient in computation. The best-performing strategy is calculating the entropy value for each point based on the predictions from multiple anchors, which surpassed random sampling by $0.56\%$, $0.92\%$ and $0.80\%$ mAP at the first three steps. Surprisingly, the result at the second step ($\mathcal{P}_2$) even outperformed the random sampling counterparts with larger annotation costs and longer training time ($\mathcal{P}_3$). Unless otherwise specified, we use \textit{Entropy} to denote this best-performing strategy in the rest of this paper.

For the \textit{Entropy} strategy mentioned above, we always chose the most uncertain point of each instance for labeling, and we were also curious about the inverse situation that the most certain points (\ie, point with the lowest entropy value) were selected instead. As shown in Fig.~\ref{fig:ablation_selection_strategy} (black dashed line), the points with the lowest entropy even performed worse than the randomly sampled points, which verified that the proposed entropy metric is a simple yet effective way to estimate the point informativeness.

\noindent\textbf{Fine-Tuning Schedule.} We compared different fine-tuning schedules at each active learning step. For the 10k schedule ($\sim$1.3 epoch), the learning rate was fixed at $0.0001$. For the 90k schedule (12 epochs), the initial learning rate was $0.01$ and reduced by a factor of 10 at iteration 60k and 80k, respectively. As shown in Fig.~\ref{fig:ablation_training_schedule}, we observed that the longer the training time, the larger the gaps between active point selection and random sampling. When adopting the 90k schedule, the gap reached to $+1.77\%$ at the third step ($\mathcal{P}_3$). The results suggest that the actively acquired point is indeed more informative, and the longer training time can further release its potential. In this paper, unless otherwise specified, the 30k schedule was used in experiments for convenience.

\begin{figure}[t]
  \centering
  \begin{subfigure}[t]{0.49\textwidth}
    {\includegraphics[width=\textwidth]{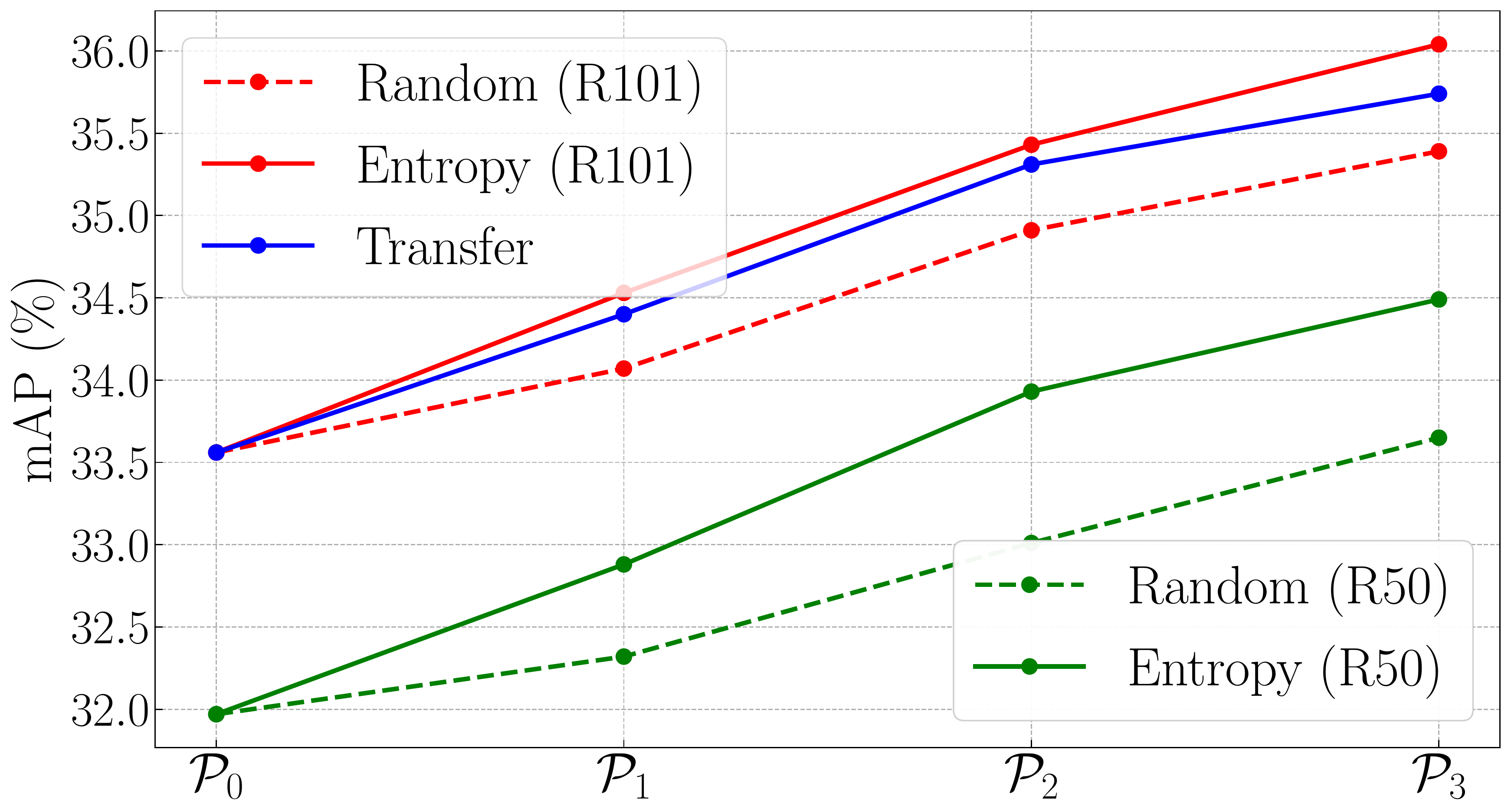}}
    \caption{CondInst with ResNet-101}
    \label{fig:ablation_r101}
  \end{subfigure}
  \hfill
  \begin{subfigure}[t]{0.49\textwidth}
    {\includegraphics[width=\textwidth]{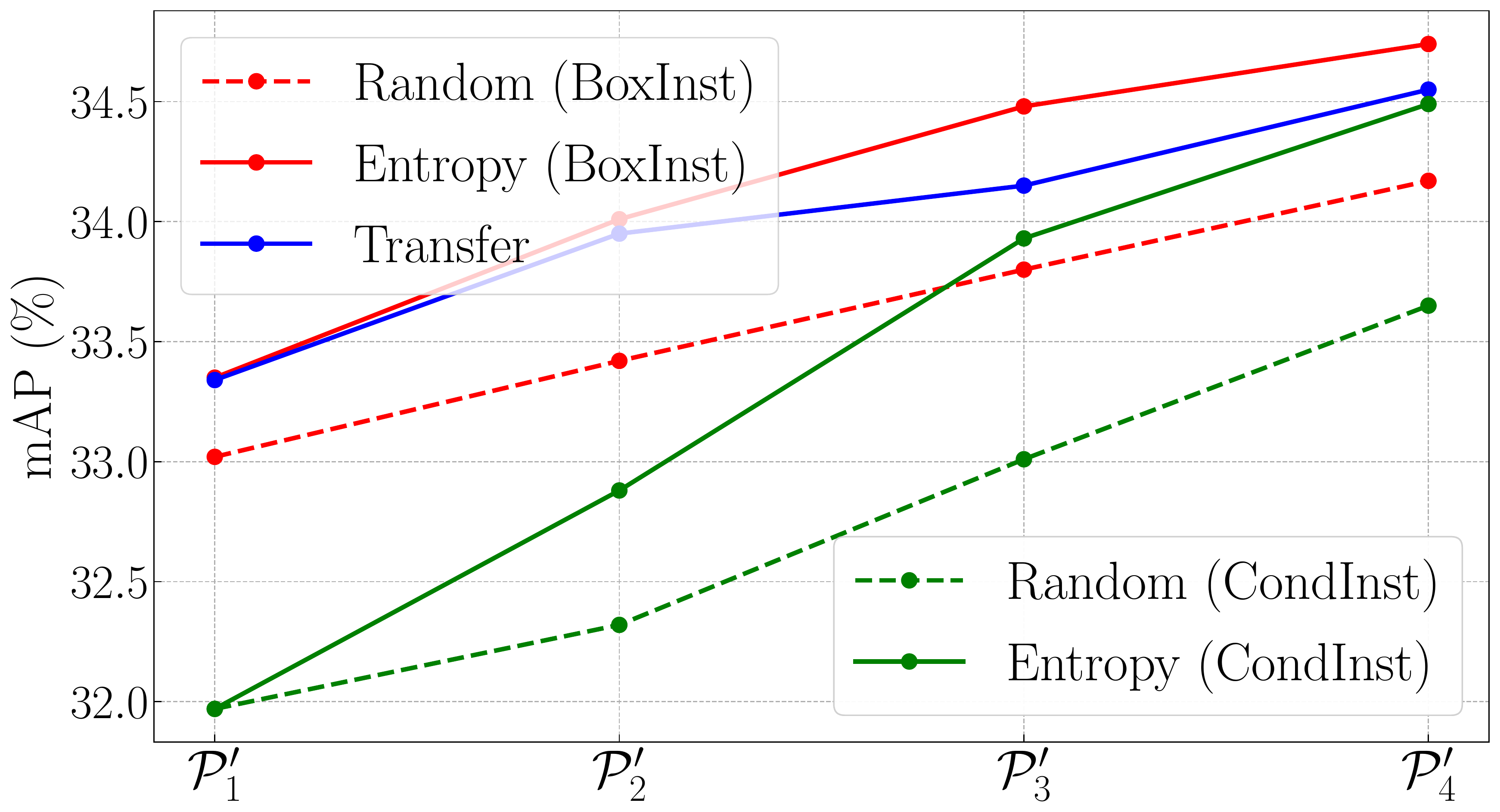}}
    \caption{BoxInst~\cite{boxinst} with ResNet-50}
    \label{fig:ablation_boxinst}
  \end{subfigure} \\
  \caption{The results of taking CondInst with ResNet-101 and BoxInst~\cite{boxinst} as the instance segmentation model, respectively. \textit{Transfer} indicates that the points are acquired from CondInst with ResNet-50 and transferred to this model.}
  \label{fig:ablation_model}
\end{figure}

\noindent\textbf{Instance Segmentation Model.} In our experiments, we mainly used the CondInst with ResNet-50 backbone as the instance segmentation model. Actually, the proposed APIS setting can be studied with a diverse set of models, and we also studied two alternatives. (a) \textit{Larger backbone}: Using the higher-capacity ResNet-101 as the backbone, the similar observation can be made that active selection works consistently better than random sampling, as shown in Fig.~\ref{fig:ablation_r101}. (b) \textit{Boxly-supervised model}: Recall that $\mathcal{P}_0$ in above experiments is a set of randomly sampled points which is used to initialize the model to obtain the initial mask predictions. A number of works~\cite{hsu2019weakly,khoreva2017simple,lan2021discobox,boxinst} attempted to predict masks by training with box annotations only. By adopting these models as the initial model in APIS, we can eliminate the need of $\mathcal{P}_0$. In Fig.~\ref{fig:ablation_boxinst}, we adopted BoxInst~\cite{boxinst}, a dominant method in the boxly-supervised instance segmentation area, as the initial model, and each instance was labeled with $s$ points in $\mathcal{P}\sp{\prime}_s$ (in comparison, $s+1$ in $\mathcal{P}_s$). As shown, the proposed strategy also worked in this case. With the power of BoxInst, similar or even better results were achieved with fewer points, \eg, the result of $\mathcal{P}\sp{\prime}_1$ surpassed $\mathcal{P}_1$ (in Fig.~\ref{fig:ablation_selection_strategy}) by $0.47\%$ mAP, although $\mathcal{P}_1$ has more points.

\noindent\textbf{Transferability.} We studied the transferability of the actively acquired points. As shown in Fig.~\ref{fig:ablation_model} (blue lines), we transferred the points acquired from one model (CondInst with ResNet-50) to other models, CondInst with ResNet-101 and BoxInst, respectively. Specifically, the point set $\mathcal{P}_s$ of the former model served as the supervision for the latter two models in the $s^\mathrm{th}$ step. As shown, the results of the transferred points are slightly lower than the results of selecting points for those models from scratch, but still higher than random sampling.

\begin{figure}[t]
  \centering
  \begin{minipage}[t]{.49\textwidth}
    \subcaptionbox{Visualization}
      {\includegraphics[width=\textwidth]{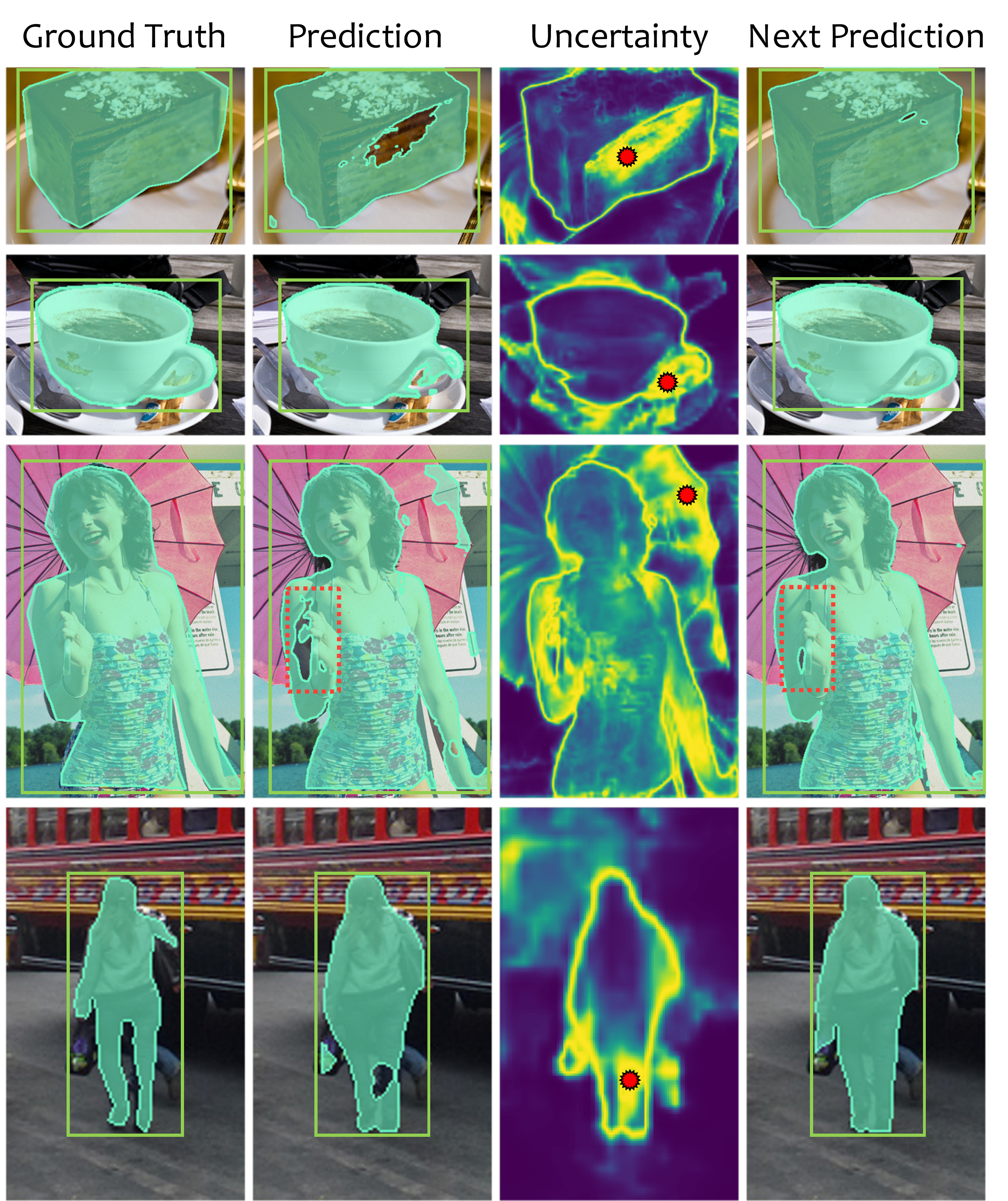}}
  \end{minipage}
  \begin{minipage}[b]{.465\textwidth}
    \subcaptionbox{Point accuracy curves}
      {\includegraphics[width=1.0\textwidth]{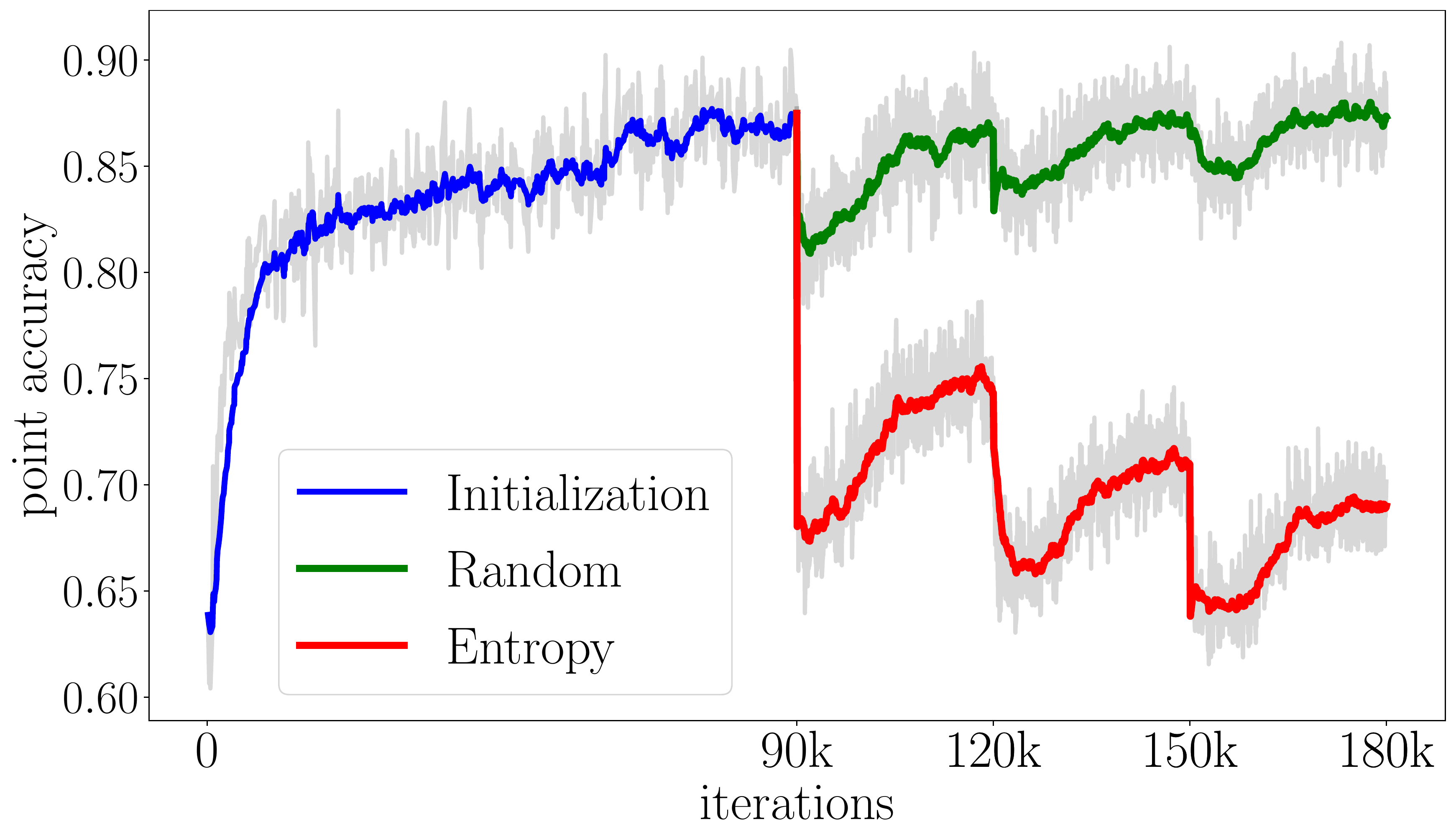}}
    \subcaptionbox{Point distribution}
      {\includegraphics[width=1.0\textwidth]{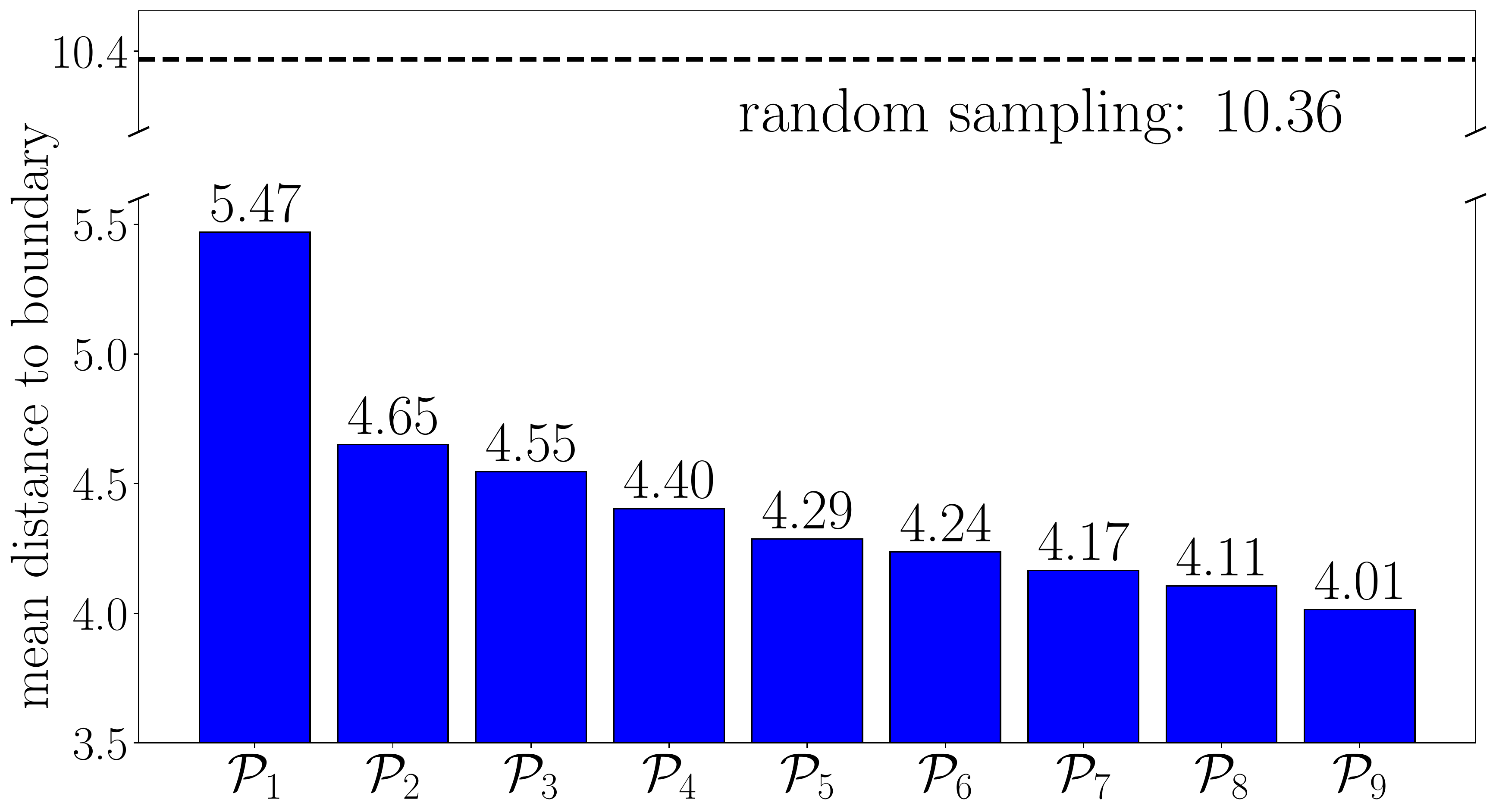}}
  \end{minipage}
  \caption{
      (a) Visualization of (from left to right): ground-truth masks, mask predictions, uncertainty maps, and mask predictions after fine-tuning with the selected points (red spots) for some instances.
      (b) Accuracy curves of the actively acquired points and random points during training.
      (c) The mean distances to the object boundaries of the actively acquired points for each step. The mean distance for random points is provided for reference (dash line).
      \label{fig:analyses}
  }
\end{figure}

\subsection{Analysis for APIS} \label{sec:analysis}

\textbf{Visualization Analysis.} To better understand how the point is selected and how it works, we visualized the mask predictions, uncertainty maps, and the selected points for some instances, as shown in Fig.~\ref{fig:analyses}a. We observed that the highlighted regions in uncertainty maps often corresponded to the two types of mistakes in the mask predictions. One is the typical over-segmented or under-segmented regions (\eg, holes on the object, or patches on the background). Another one is the error around the boundaries of the predicted masks (\ie, imprecise boundaries). Therefore, the points selected from these misclassified regions can provide a valuable feedback about how the model performs and it is possible to correct the mistakes by fine-tuning with the selected points in the subsequent active learning steps (\eg, the last column). Interestingly, some prediction errors were still corrected even without sampling the points from those regions (\eg, dashed boxes in the $3^\mathrm{rd}$ row). The reason might be that the patterns of these error regions also appeared in other instances that have the desired annotations. On the other side, there were also some failure cases where the prediction around the selected point even got worse after fine-tuning (\eg, the $4^\mathrm{th}$ row). More visualized results are included in Appendix~\ref{appendix_APIS}.

\noindent\textbf{Point Accuracy Curves.} In addition to the case studies above, the accuracy (\ie, accuracy of binary classification for points during training) curves of the actively acquired points and randomly sampled points are plotted in Fig.~\ref{fig:analyses}b. As shown, the accuracy of the actively acquired points dropped dramatically at the beginning of each step (iteration 90k, 120k and 150k, respectively), which shows that the selected points were often misclassified at the previous step and it is possible to correct the errors by fine-tuning the model with their labels (\eg, the first three rows of Fig.~\ref{fig:analyses}a). In contrast, the accuracy of random points dropped slightly at each time and always stayed at a high level, which indicates that most of these points were already handled by the model and of course less informative. In addition, the accuracy improvements usually decreased with more steps. It suggests that, with more steps, the mask prediction gets gradually better and the selected point usually gets harder.

\noindent\textbf{Point Distribution.} We calculated the distance between the actively acquired points and the ground-truth instance boundaries, and show the mean distances at each step in Fig.~\ref{fig:analyses}c. As shown, the selected points were usually closer to boundaries than the random points and the mean distance became smaller with more steps. This is as expected --- as the model gets gradually better (\eg, over/under-segmented regions has been corrected), the predicted boundaries get closer to the actual object boundaries, thus the remaining high-entropy points are mostly located around object boundaries (see the $2^\mathrm{nd}$ column of Fig.~\ref{fig:analyses}a). However, the points around object boundaries are inherently hard to classify even with full supervisions, as studied in previous works~\cite{tang2021look,refinemask}, and their labels might be noisy due to the coarse polygon-based mask annotations in MS-COCO. In summary, with more steps, the algorithm tends to select points around object boundaries, yet these annotations, though being difficult, often bring marginal performance gain, which confirmed the observations in above analysis.

\noindent\textbf{Point Difficulty.} From the accuracy decrements (after adding new points) in Fig.~\ref{fig:analyses}b we can calculated that about $51\%$ of the actively acquired points ($\mathcal{P}_1$) were misclassified by the previous model ($\mathcal{M}_0$), while the ratio is $23\%$ for random sampling, which suggests that the actively acquired points are more difficult for the model to learn. To study the influence of point difficulty, two experiments were conducted where we selected the points with maximum error or with minimum error at each step.
The results were compared in terms of both mask mAP (Table~\ref{table:result_error_point}) and point accuracy (Fig.~\ref{fig:point_accu_error_point}). For \textit{Least Error}, although the point accuracy (orange line) always stayed at a high level, the results of mAP still lagged behind random sampling, which suggests that these well-classified points were too easy for the model and of course less informative. For \textit{Max Error}, the results of mAP were extremely poor while the point accuracy (black line) still increased during each training step. The reason might be that these points were mostly the hard cases and training with them directly made the model overfitted.
Two conclusions can be drawn from the above results:
(a) Point difficulty can heavily impact the performance of active learning, where neither the easiest nor the most difficult points should be selected. The proposed \textit{Entropy} metric achieves a reasonable balance in this factor.
(b) APIS is a challenging but non-trivial problem. It is still difficult to determine which point has a potentially large contribution even provided the mask labels.
The results and analyses suggest that uncertainty is a better solution towards the right direction than correctness.

\setlength{\tabcolsep}{2pt}
\begin{table}[t]
\begin{minipage}[b]{0.48\linewidth}
    \centering
    \captionof{table}{Sampling points with different difficulties. $\star$ indicates that mask labels were used for selection.}
    \footnotesize
    \begin{tabular}{|c|cccc|}
      \hline
			Strategy & $\mathcal{P}_0$ & $\mathcal{P}_1$ & $\mathcal{P}_2$ & $\mathcal{P}_3$ \\
			\hline\hline
      Random & 31.97 & 32.32 & 33.01 & 33.69 \\ \hline
      Entropy & 31.97 & \textbf{32.88} & \textbf{33.93} & \textbf{34.49} \\
      $^\star$Max Error & 31.97 & 7.95 & 12.45 & 14.24 \\
      $^\star$Least Error & 31.97 & 32.16 & 32.50 & 32.95 \\
      \hline
		\end{tabular}
    \label{table:result_error_point}
\end{minipage}
\hfill
\begin{minipage}[b]{0.5\linewidth}
    \centering
    \includegraphics[width=1.0\linewidth]{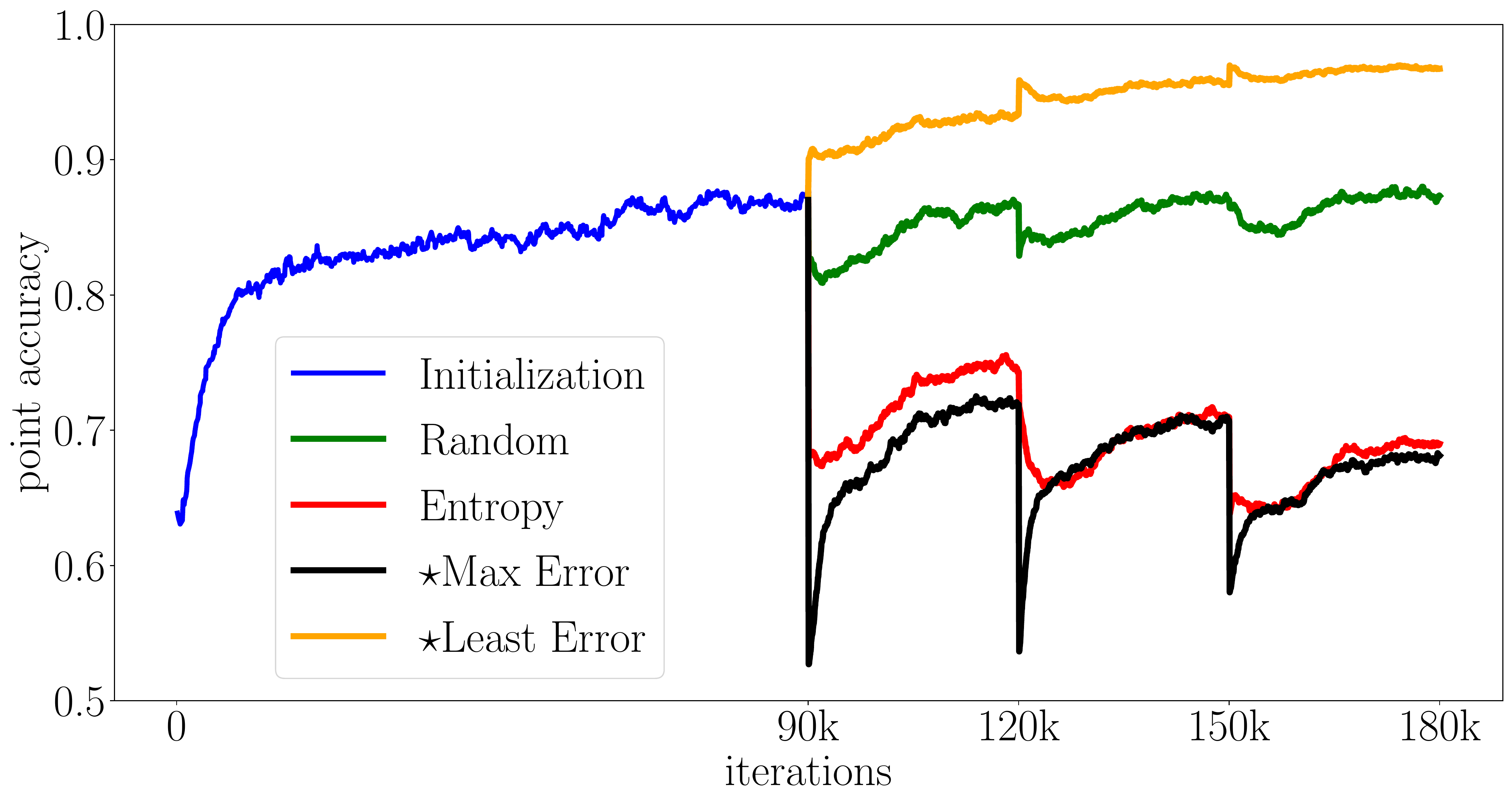}
\end{minipage}
\begin{minipage}{\linewidth}
    \captionof{figure}{The point accuracy (measured on \texttt{train2017}) curves of sampling points with different difficulties. At the first step ($\mathcal{P}_1$), the misclassification ratio was $82\%$ and $7\%$ for \textit{Max Error} and \textit{Least Error}, respectively.}
    \label{fig:point_accu_error_point}
\end{minipage}
\end{table}

\subsection{Comparison to Other Learning Strategies}
\textbf{Comparison of APIS and AFIS.} As introduced in Sec.~\ref{sec:baseline}, a baseline setting AFIS was established for comparison. For fair comparison, the annotation budget and training time should be the same with APIS during each step. As stated in previous works~\cite{pointsup,mscoco}, it takes $0.9$ seconds on average to label a point and 79.2 seconds to create a polygon-based instance mask in MS-COCO. In above experiments, one point was labeled for each instance at each step, thus 860000 points in total. If the same budget is allocated to instances, we can annotate masks for $860000/(79.2/0.9)\approx9773$ instances. Unlike previous works that the annotation cost for different images are treated identical, in our case the cost is proportional to the number of instances in the image, which varies across images. Therefore, given a fixed annotation budget, the number of annotated images is dependent on the sampling strategy.

\begin{figure}[t]
  \centering
  \begin{subfigure}[b]{0.49\textwidth}
    {\includegraphics[width=\textwidth]{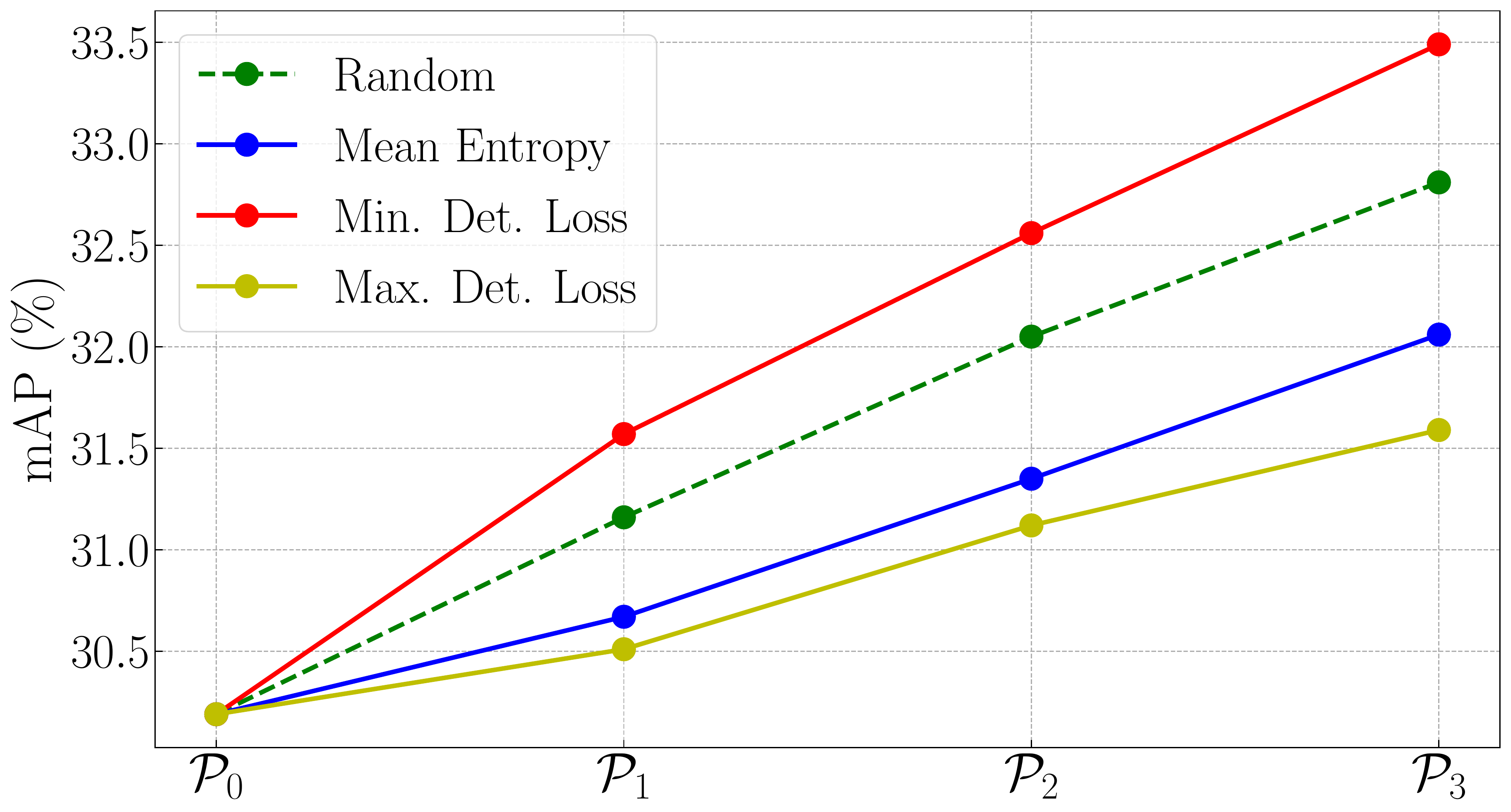}}
    \caption{AFIS (image-level)}
    \label{fig:AFIS_image_level}
  \end{subfigure}
  \hfill
  \begin{subfigure}[b]{0.49\textwidth}
    {\includegraphics[width=\textwidth]{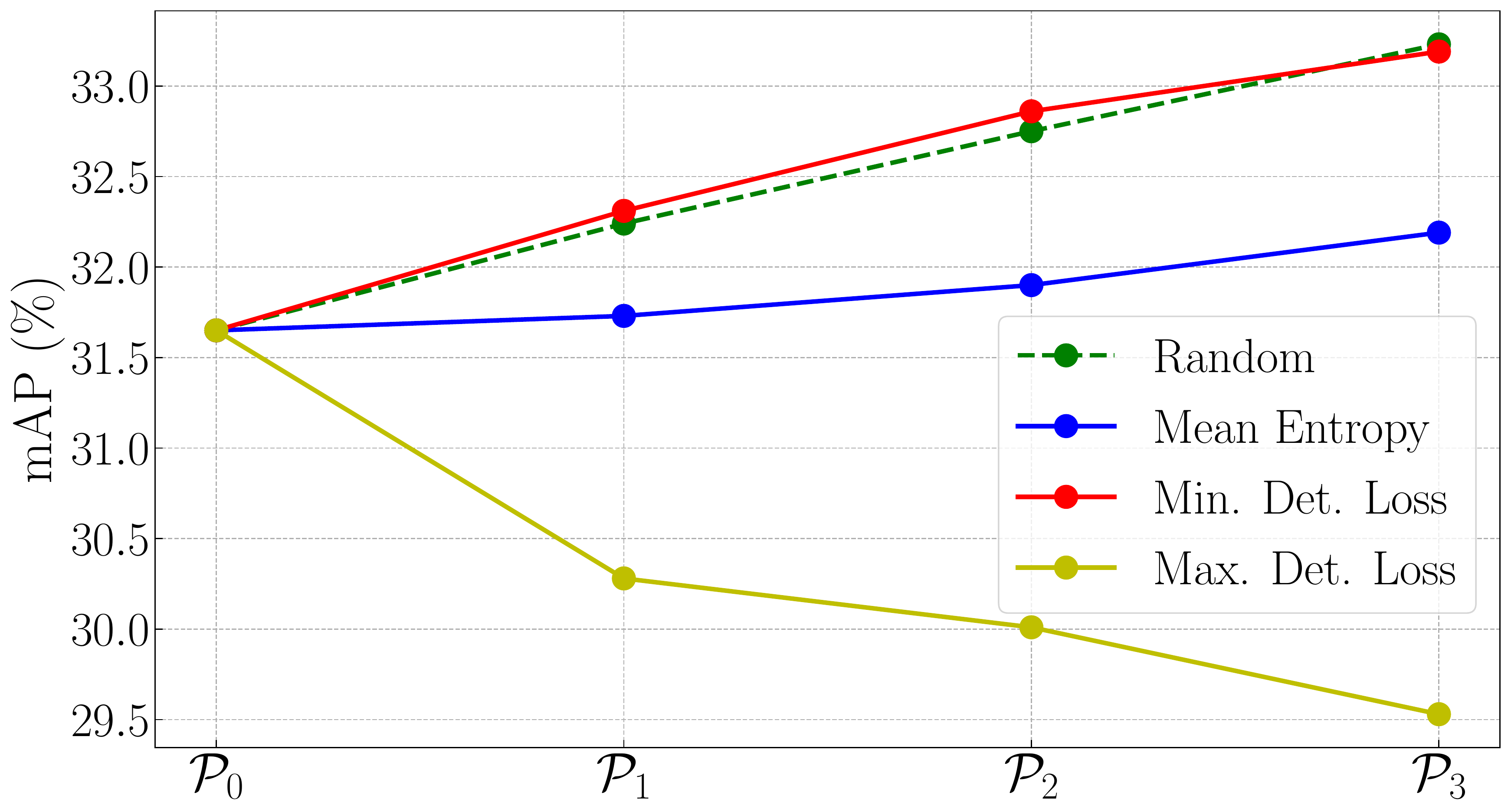}}
    \caption{AFIS (instance-level)}
    \label{fig:AFIS_instance_level}
  \end{subfigure} \\
  \caption{Comparison of different image-level and instance-level selection strategies designed for AFIS. See Sec.~\ref{sec:baseline} and Appendix~\ref{appendix_AFIS} for details.}
  \label{fig:ablation_AFIS}
\end{figure}

We compared different sampling strategies for the case of \textit{image-level} selection and \textit{instance-level} selection, respectively. $\mathcal{P}_s$ here indicates that the cost for labeling images or instances is exactly the same as $\mathcal{P}_s$ in APIS. Similarly, some images or instances were randomly selected (with the same budget as $\mathcal{P}_0$) for model initialization.
As shown in Fig.~\ref{fig:ablation_AFIS}, the results of the \textit{Mean Entropy} strategy were unsatisfactory in both cases, even lagging behind random sampling. On the other hand, the strategy that selecting instances with the lowest detection loss (\textit{Min. Det. Loss}) usually produced on par or even better results than other competitors. Since AFIS is not the focus of this paper, we provided more results and analyses of AFIS in Appendix~\ref{appendix_AFIS}.

The best-performing strategies found above were adopted for comparison in Fig.~\ref{fig:mAP_curve}. Three conclusions can be drawn from the results:
(1) Point-based supervision is an effective way for training instance segmentation models. Even with random points, the model can still outperformed those trained with mask supervisions under the same annotation costs and training time.
(2) Active learning is a label-efficient training strategy for instance segmentation, especially when the annotation budget is limited. As shown, the active selection strategy usually performed better than random sampling in all settings.
(3) The proposed APIS setting, combining active learning and point-based supervision, is a more powerful yet economic choice to train instance segmentation models under limited annotation budgets. The model developed under this setting consistently outperformed all other competitors at each active learning step. For example, at the $5^\mathrm{th}$ step ($\mathcal{P}_5$), our model achieved on par or even better results than other models with more points and longer training time (\eg, $\mathcal{P}_9$).

\setlength{\tabcolsep}{3pt}
\begin{table}[t]
\begin{minipage}[b]{.49\linewidth}
  \centering
  \includegraphics[width=1.0\linewidth]{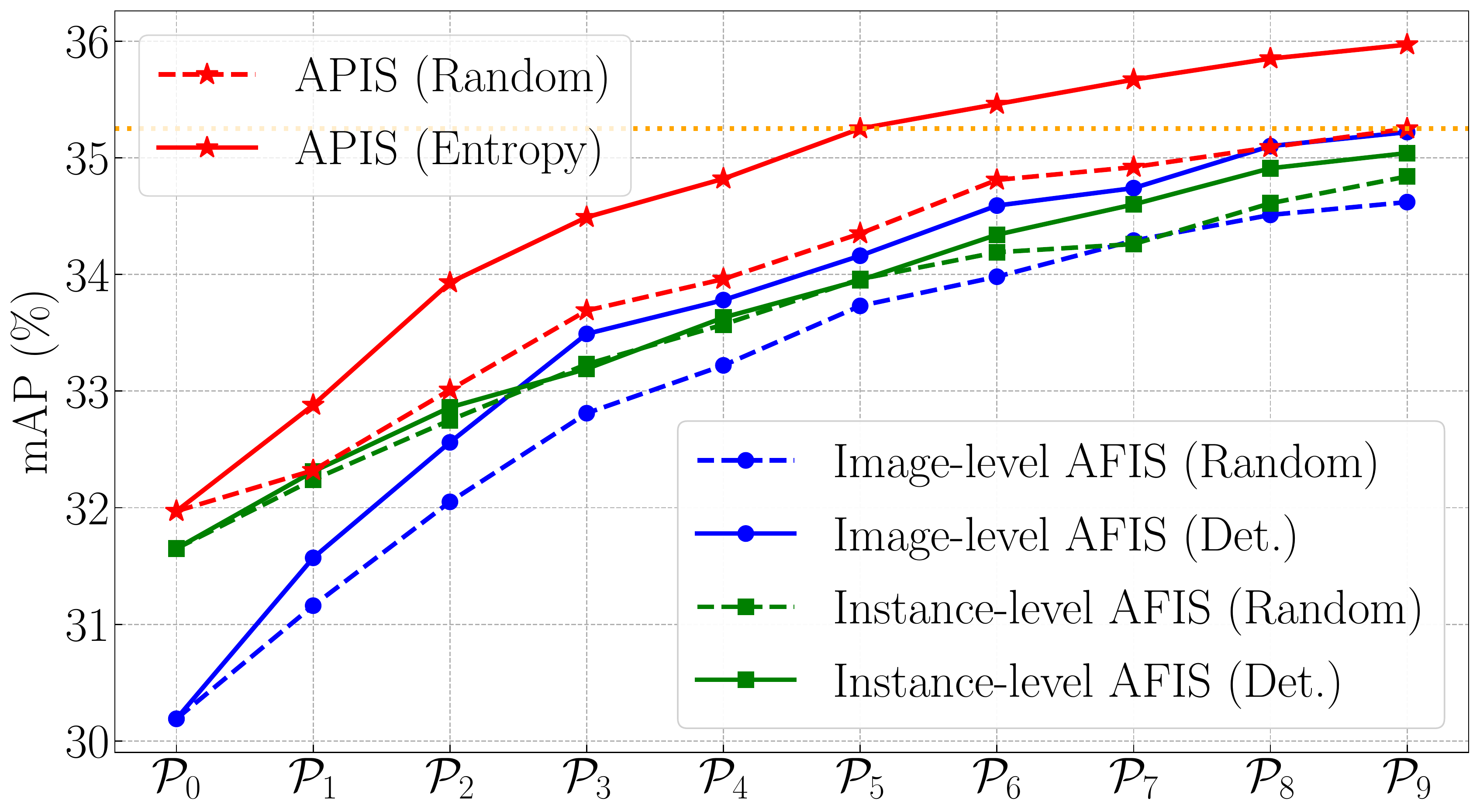}
  \captionof{figure}{Comparison of APIS and AFIS with the same annotation budget and training time. Det. indicates the \textit{Min. Det. Loss} strategy.}
  \label{fig:mAP_curve}
\end{minipage}
\hfill
\begin{minipage}[b]{0.49\linewidth}
    \centering
    \captionof{table}{Comparison of APIS and weakly-supervised instance segmentation methods (with ResNet-50 backbone). $\dagger$: our impl. (no point aug.).}
    \vspace{3mm}
    \renewcommand{\arraystretch}{1.15}
    \begin{tabular}{|l|c|c|c|}
      \hline
      Method &Anno. &Iter. &mAP \\
      \hline\hline
      CondInst~\cite{condinst} &fully sup. &270k &37.5 \\ \hline
      DiscoBox~\cite{lan2021discobox} &$\{\mathcal{C},\mathcal{B}\}$ &270k &31.4 \\
      BoxInst~\cite{boxinst} &$\{\mathcal{C},\mathcal{B}\}$ &270k &31.8 \\ \hline
      PointSup$^{\dagger}$~\cite{pointsup} &$\{\mathcal{C},\mathcal{B},\mathcal{P}_{10}\}$ &270k &35.4 \\
      APIS (ours) &$\{\mathcal{C},\mathcal{B},\mathcal{P}_7\}$ &270k &35.4 \\
      APIS (ours) &$\{\mathcal{C},\mathcal{B},\mathcal{P}_{10}\}$ &360k &36.0 \\
      \hline
    \end{tabular}
    \label{table:weakly_supervised}
\end{minipage}
\end{table}

\textbf{Comparison to Weakly-Supervised Methods.} In Table~\ref{table:weakly_supervised}, we compared APIS with some existing weakly-supervised instance segmentation methods. Compared to boxly-supervised methods~\cite{lan2021discobox,boxinst}, labeling points additionally usually leads to considerable improvements. Compared to PointSup~\cite{pointsup} (based on CondInst with ResNet50) trained with 10 randomly sampled points, APIS achieved the same results with 3 fewer points per instance. We argue that random sampling not considered the informativeness of different points, thus leading to sub-optimal results. In spite of this, we have to realize that APIS and PointSup varied significantly in motivation, which optimized for active learning and weakly-supervised learning, respectively.

\section{Conclusion}
In this paper, we propose APIS, a new active learning setting for instance segmentation where the most informative points are selected for annotation. We formulate this setting and propose several sampling strategies. Extensive experiments and detailed analysis on MS-COCO demonstrate that APIS is a powerful but economic learning strategy for training instance segmentation models with limited annotation budgets. We hope this work could inspire future researches on related topics, \eg, point-based supervision and label-efficient learning.

\noindent\textbf{Acknowledgements.} This work was supported in part by the National Natural Science Foundation of China (Nos. U19B2034, 62061136001 and 61836014).

%
%
\bibliographystyle{splncs04}
\bibliography{egbib}

\appendix

\section*{Appendix}

\setcounter{table}{0}
\setcounter{figure}{0}
\setcounter{equation}{0}
\renewcommand{\theequation}{S\arabic{equation}}
\renewcommand{\thefigure}{S\arabic{figure}}
\renewcommand{\thetable}{S\arabic{table}}
\renewcommand{\theHfigure}{S\arabic{figure}}
\renewcommand{\theHtable}{S\arabic{table}}

\section{More Details and Results of AFIS} \label{appendix_AFIS}

Since there is no existing work can be directly compared to APIS, we established the baseline setting \textit{active fully-supervised instance segmentation (AFIS)}. The mask annotations (image-level or instance-level) are queried by the model during each active learning step, which is conceptually similar to some existing active learning algorithm designed for image classification or object detection. In this section, we provide more description and results of AFIS.

\begin{figure*}[t]
\begin{center}
  \includegraphics[width=\linewidth]{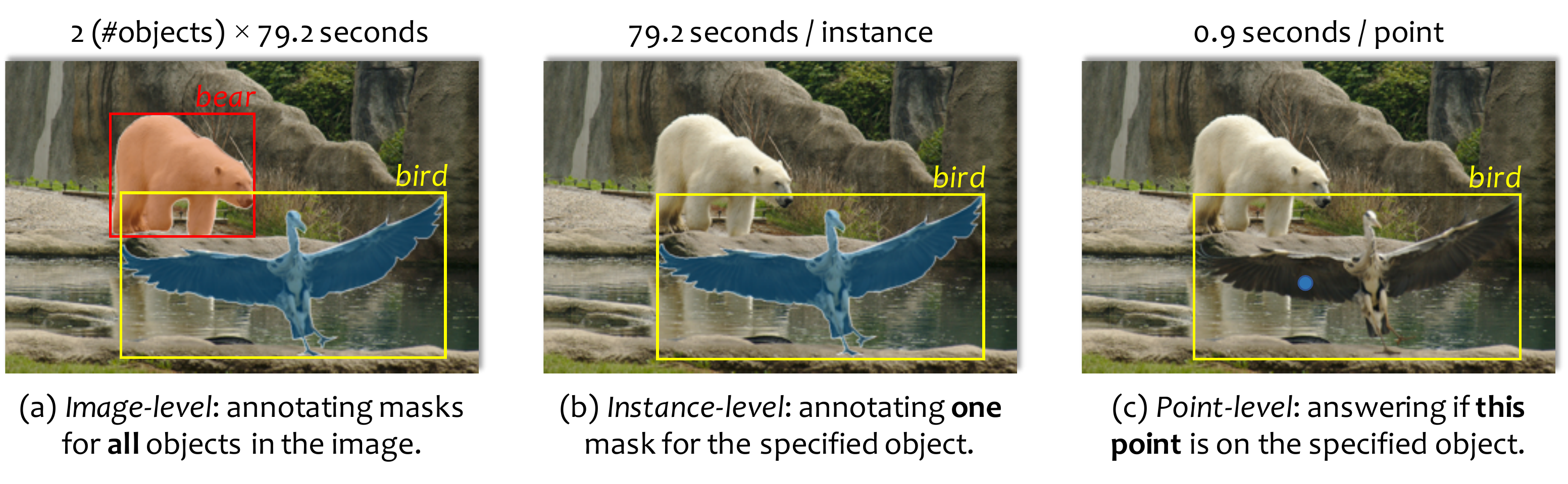}
\end{center}
\vspace{-8mm}
  \caption{
    Illustration of different annotation schemes (\textit{image-level}, \textit{instance-level}, and \textit{point-level}) from the perspective of human annotators. The averaged annotation time for these schemes is provided for reference (values adopted from~\cite{pointsup}).
  }
  \label{fig:annotation_style}
\end{figure*}

\textbf{Annotation Schemes.} Fig.~\ref{fig:annotation_style} illustrates different annotation schemes from the perspective of human annotators, as well as the corresponding approximate annotation time. Compared to AFIS, APIS can be studied in a more fine-grained manner because it allocates annotation budgets to pixels, and the annotation of points is considerably faster and cheaper.

\textbf{Sampling Strategy of AFIS.} In Fig.~\ref{fig:ablation_AFIS}, we compared different sampling strategies for the case of \textit{image-level} selection and \textit{instance-level} selection, respectively.

\textbf{Firstly}, the results of the \textit{Mean Entropy} strategy were unsatisfactory in both cases, even lagging behind the results of random sampling. We diagnosed the problem and found the main reason is that the instances selected under this metric are usually small objects. For example, over $76\%$ of the selected instances at the first step are small (\ie, area$\textless32^2$, as defined in COCO). For the larger objects, there were usually a lot of low-entropy points (\eg, points on the smooth interior areas or background), which decreased the mean entropy value. It is well-known that the the model's performance is usually poor on small objects even with the mask supervision~\cite{singh2018analysis}, thus the annotated instances in our cases were not so effective.
\textbf{Secondly}, we used the detection loss (\eg, GIoU Loss) to measure the \textit{Detection Quality}, \ie, the lower the loss, the higher the quality. As shown, the strategy that selecting with the lowest detection loss (\textit{Min. Det. Loss}) usually produced better results than random sampling for the image-level selection, while for the instance-level selection, the performance was usually on par with random sampling. In addition, we also studied the opposite strategy that selecting with the highest detection loss (\textit{Max. Det. Loss}) and found the results are poor. The results suggest that instance-level cues (\eg, detection quality) are somehow important for AFIS. From these results, we conjecture that instance-level cues (\eg, detection quality) may provide complementary information to assist APIS or a mixed setting that both instance-level and point-level supervision can be chosen. Although the strategy for AFIS has not been thoroughly optimized, we believe that it is sufficient to serve as a reasonable baseline for APIS and provides some useful guidance for future researches in this area.

\begin{figure*}[t]
\begin{center}
  \includegraphics[width=\linewidth]{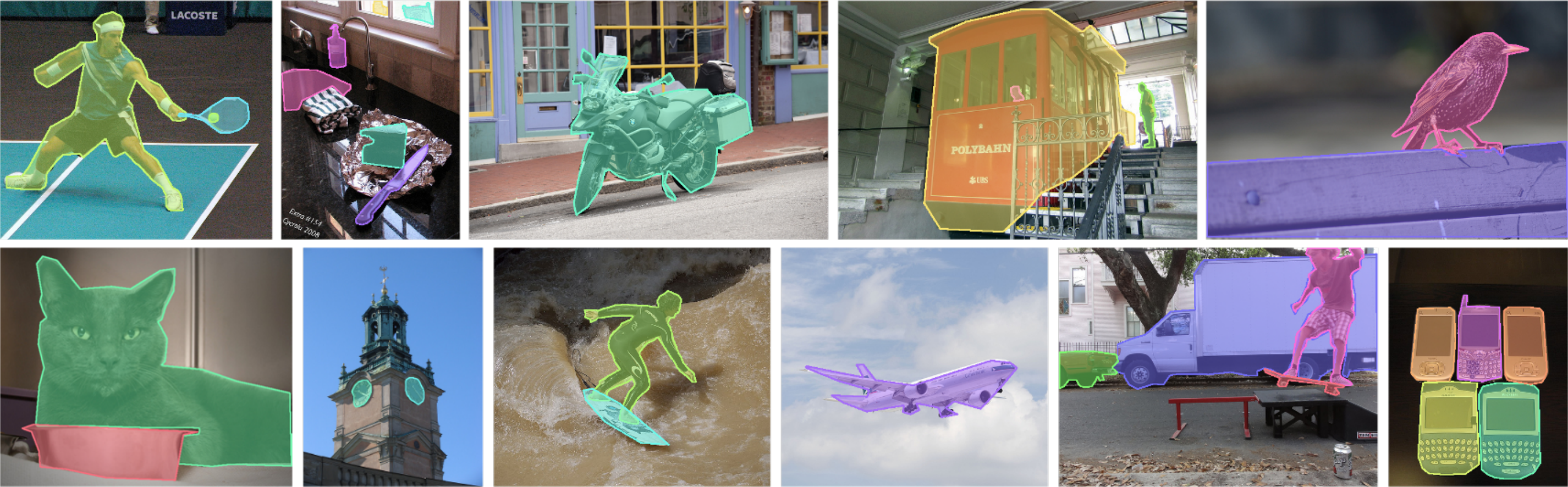}
\end{center}
\vspace{-6mm}
  \caption{
    Examples of the selected images under the \textit{Detection Quality} metric for \textit{image-level} AFIS.
  }
  \label{fig:image_level_examples}
\end{figure*}

\begin{figure*}[t]
\begin{center}
  \includegraphics[width=\linewidth]{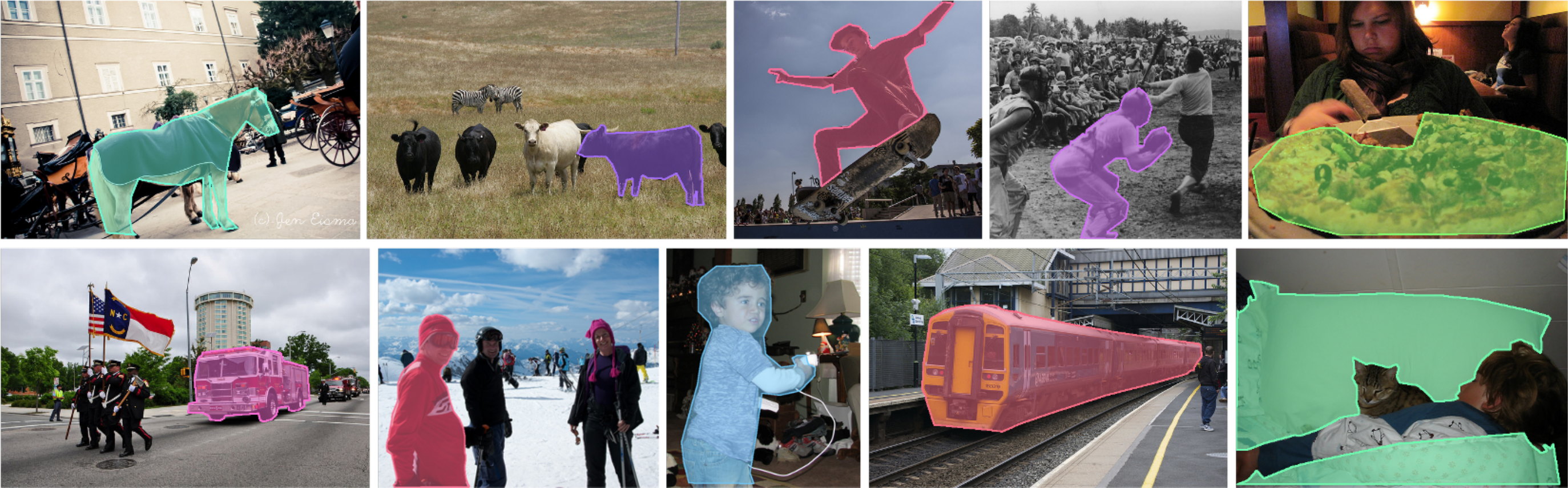}
\end{center}
\vspace{-6mm}
  \caption{
    Examples of the selected instances under the \textit{Detection Quality} metric for \textit{instance-level} AFIS.
  }
  \label{fig:instance_level_examples}
\end{figure*}

\textbf{Qualitative results of image-level AFIS.} Fig.~\ref{fig:image_level_examples} shows some examples of the selected images under the \textit{Detection Quality} (\ie, minimum detection loss) metric. As shown, the selected image usually contains fewer objects. Empirically, if the annotation budget, the same as labeling one point for each instance at an active learning step, was allocated to images, we can annotate masks for about $3435$ images. Each image contains $2.8$ objects on average, which is considerably fewer than the number of annotated objects per image ($7.7$) in MS-COCO. This observation is contrary to most works on active object detection where the algorithms usually prefer the images with more objects~\cite{haussmann2020scalable} since the annotation costs for different images are considered the same in their experiments. In our setting, the annotation cost is proportional to the number of objects in the image, which is closer to the real-world scenarios.

\textbf{Qualitative results of instance-level AFIS.} Fig.~\ref{fig:instance_level_examples} shows some examples of the selected instances under the \textit{Detection Quality} metric. We found that the selected instances usually covered a large area, \eg, over $63\%$ of the selected instances at the first step were large object (\ie, area$\textgreater96^2$ pixels, as defined in COCO), which is consistent with the observation that the detection results of larger objects are usually better than the results of smaller objects.

\section{More Results of APIS} \label{appendix_APIS}

\hspace{\parindent}\textbf{Relation to Object Scale.} We empirically found that the model trained with actively acquired points performed much better on the larger instances (higher $\Delta$AP$_{L}$) than random points, as listed in Tabel~\ref{table:AP_area}. It indicates that for larger instances, the informativeness of different points is more diverse than smaller one. Besides, the smaller instances are inherently hard to recognize no matter which type of label is given.

\textbf{APIS on Cityscapes.} We additionally reported the results on Cityscapes~\cite{cityscapes}. The results (red lines in Fig.~\ref{fig:citys}) validate the same conclusion. In this study, we follow the previous work~\cite{maskrcnn} to use MS-COCO pre-training, and the results are slightly unstable due to the small dataset size.

\textbf{Box-free APIS.} In this work, we studied APIS with box-level annotations, but it is also feasible by eliminating bounding box annotations. We validated this through a preliminary solution (\textit{i.e.}, generating pseudo box labels using an off-the-shelf detector to assist APIS) on the Cityscapes dataset. As shown in Fig.~\ref{fig:citys} (the blue lines), the mask AP dropped by $1\%$--$2\%$ due to the inaccurate box labels. Note that the decrease is moderate since we used a detector that produced high-quality pseudo boxes (only to show the feasibility), and the results might be lower with low-quality pseudo box labels.
Additionally, there exist advanced point-based detectors~\cite{chen2021points} to integrate, which we leave for future work.

\setlength{\tabcolsep}{7pt}
\begin{table}[t]
  \begin{minipage}[b]{0.49\linewidth}
      \centering
      \caption{The mAP improvement of the \textit{Entropy} strategy over random sampling, as well as the results on the small, medium and large instances.}
      \label{table:AP_area}
      \renewcommand{\arraystretch}{1.3}
      \begin{tabular}{|l|ccc|}
        \hline
         & $\mathcal{P}_1$ & $\mathcal{P}_2$ & $\mathcal{P}_3$ \\
        \hline \hline
        $\Delta$AP &+0.56 &+0.92 &+0.80 \\ \hline
        $\Delta$AP$_S$ &$-$0.30 &+0.37 &+0.28 \\ \hline
        $\Delta$AP$_M$ &+0.69 &+0.93 &+0.66 \\ \hline
        $\Delta$AP$_L$ &\textbf{+1.17} &\textbf{+1.53} &\textbf{+1.33} \\
        \hline
      \end{tabular}
  \end{minipage}
\hfill
\begin{minipage}[t]{.49\linewidth}
  \centering
  \includegraphics[width=1.0\linewidth]{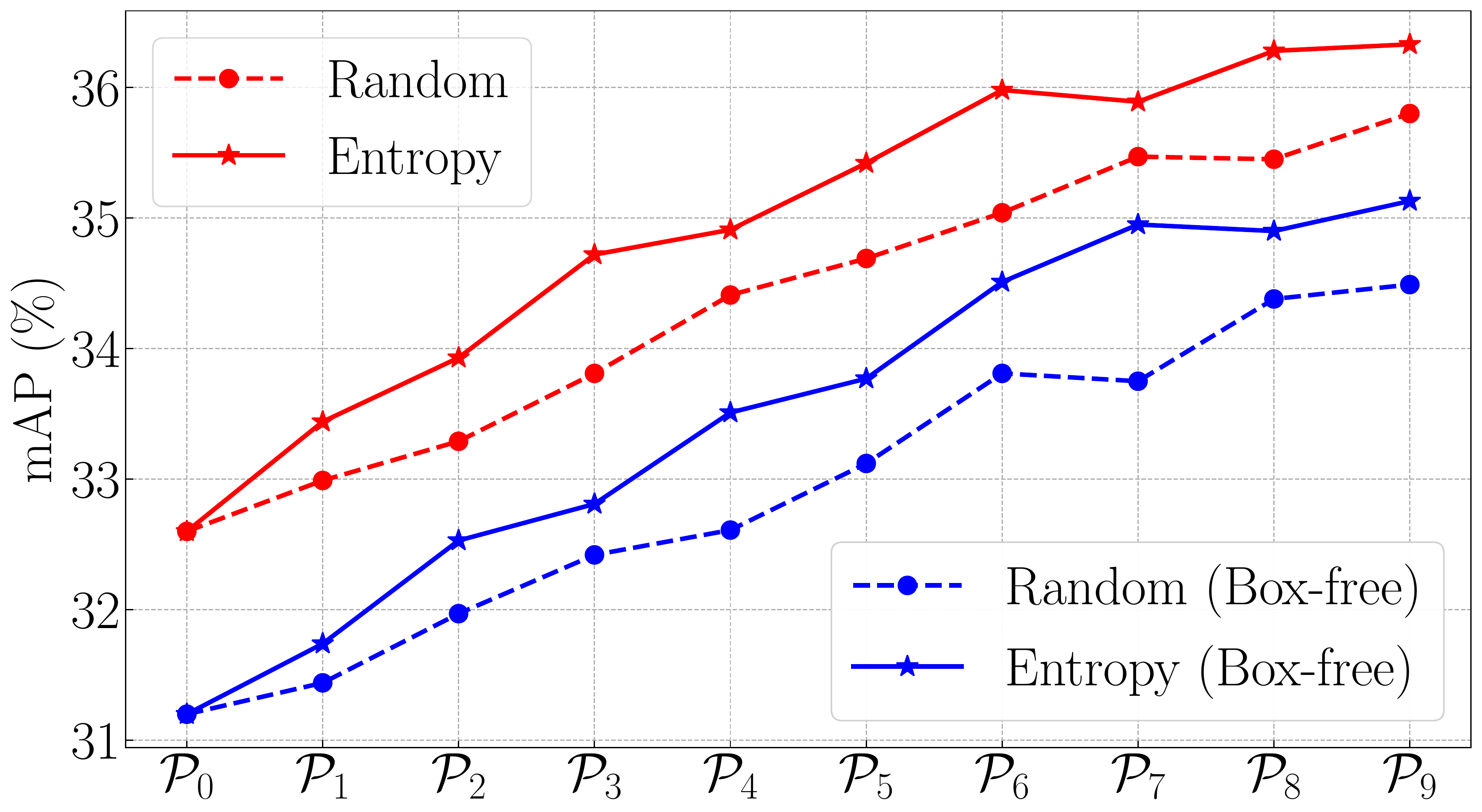}
  \captionof{figure}{APIS results on the Cityscapes dataset. Blue lines indicate the box-free APIS results.}
  \label{fig:citys}
\end{minipage}
\end{table}

\begin{figure*}[t]
\begin{center}
  \includegraphics[width=1.0\linewidth]{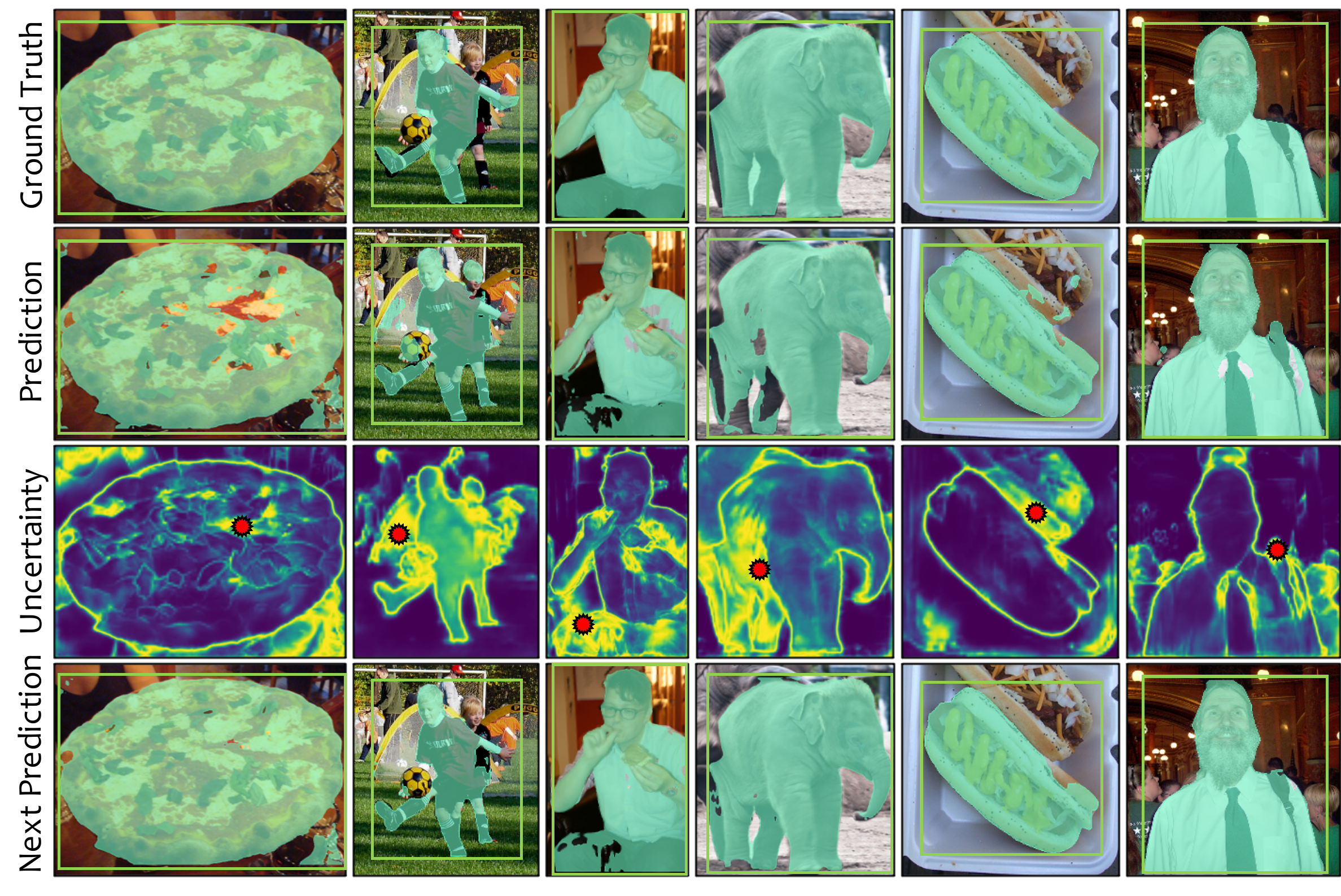}
\end{center}
  \caption{
    Visualization of (from top to bottom): ground-truth masks, mask predictions (averaged over multiple predictions), uncertainty maps (the brighter the more uncertain), and mask predictions after fine-tuning with the selected points for some instances (extension of Fig.~\ref{fig:analyses}a). The red spots indicate the selected points. Best viewed in colour.
  }
  \label{fig:more_visualization}
\end{figure*}

\textbf{More Visualization Results of APIS.} We visualized the mask predictions, uncertainty maps, and the selected points for more instances, as shown in Fig.~\ref{fig:more_visualization} (extension of Fig.~\ref{fig:analyses}a).

\section{APIS with Fewer Labeled Points} \label{appendix_fewer}
\begin{wrapfigure}{r}{0.55\textwidth}
  \centering
  \vspace{-4mm}
  \includegraphics[width=\linewidth]{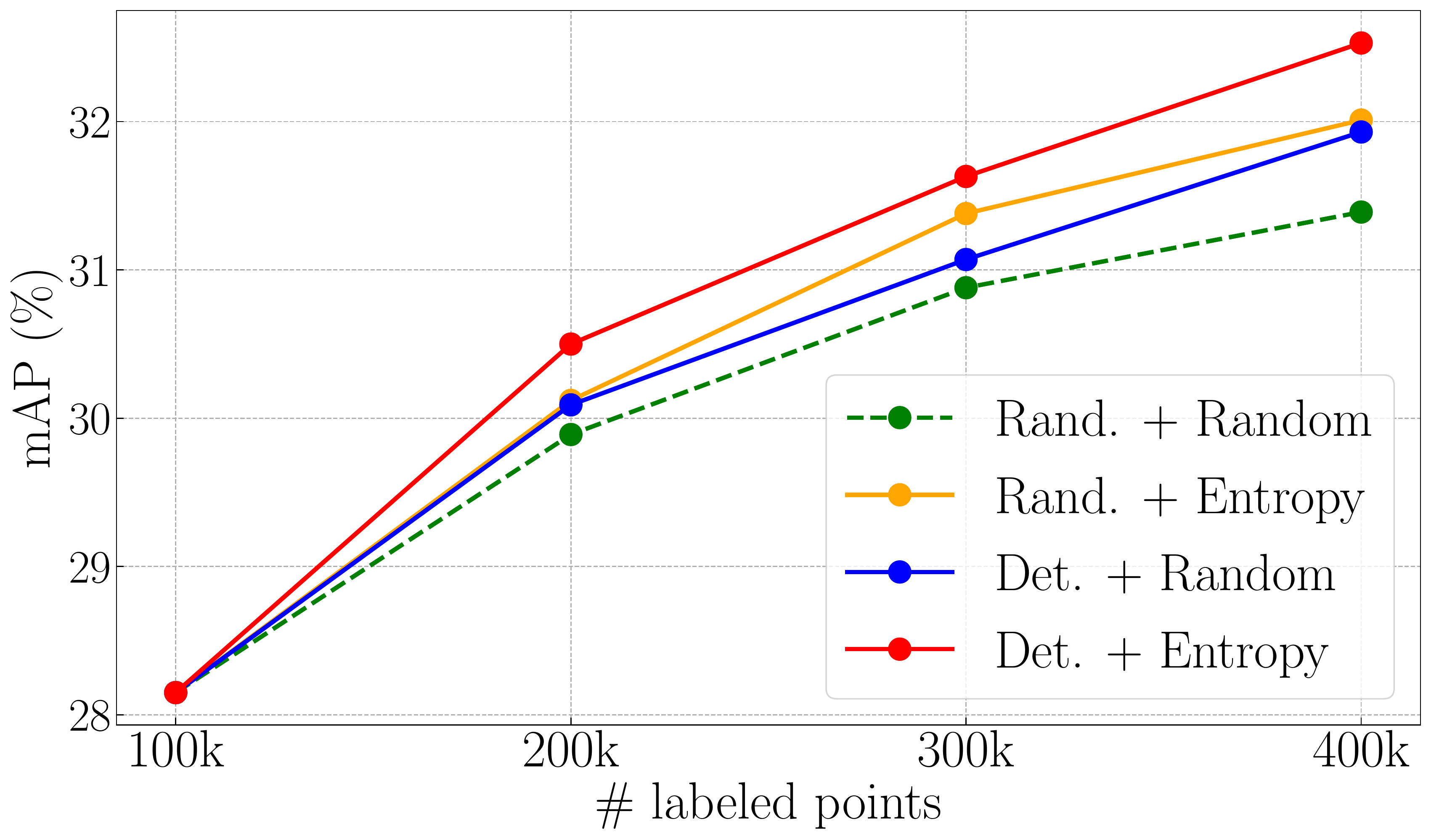}
  \caption{The results of APIS with fewer labeled points. Rand. and Det. indicate random sampling and the \textit{Min. Det. Loss} strategy, respectively.}
  \vspace{-4mm}
  \label{fig:less_point}
\end{wrapfigure}

In above experiments, we labeled one point for each instance at a step, while the difficulty of instance was not considered, \eg, some instances are easier to learn and require fewer (or even zero) points, while others may require more points. We studied this problem by reducing the annotation budget of each step to 100,000 points (8$\times$ fewer). The training pipeline keeps unchanged. We explored two different ways to select instances: random sampling and the \textit{Min. Det. Loss} strategy (similar to instance-level AFIS, see Sec.~\ref{sec:baseline}).
As shown in Fig.~\ref{fig:less_point}, the actively acquired points still worked better than random points. As for instance selection, sampling instances with higher detection quality (\ie, lowest loss) led to higher performance. With 400k points ($32.5\%$), the model outperformed the previous model trained with 860k points ($32.0\%$ for $\mathcal{P}_0$), but at the cost of longer training time ($2\times$). Both the annotation cost and computational cost should be considered when deciding the number of labeled points at each step, while the former is usually much more expensive in practice.

\end{document}